\documentclass[journal]{IEEEtran}
\renewcommand\IEEEkeywordsname{Keywords}
\usepackage{romannum}
\usepackage{amssymb}
\usepackage{amsmath,amsthm}
\usepackage{graphicx}
\usepackage{dblfloatfix} 
\usepackage{multicol}
\usepackage{lipsum}
\usepackage{algorithm}
\usepackage{algpseudocode}
\usepackage{booktabs}
\usepackage{tabularx}
\usepackage{ragged2e}
\usepackage{csquotes}
\usepackage{helvet}
\usepackage{textcomp}
\usepackage[T1]{fontenc}
\usepackage{bm}
\usepackage{verbatim} 
\usepackage{mathtools}
\usepackage{caption}
\usepackage{graphicx} 
\usepackage{threeparttable}
\usepackage{textcomp}
\usepackage{dblfloatfix}
\usepackage{tabularx}
\usepackage{color}
\usepackage{bm}
\usepackage{mathrsfs,float} 
\usepackage{epstopdf}
\usepackage{float}
\usepackage{gensymb,stmaryrd}
\usepackage{algorithmicx}
 \usepackage{enumerate}
 \usepackage{enumitem}
  \newcolumntype{P}[1]{>{\centering\arraybackslash}p{#1}}
  \newcolumntype{M}[1]{>{\centering\arraybackslash}m{#1}}

\usepackage{hyperref}
\usepackage{textcomp}
\usepackage[normalem]{ulem}
\usepackage{xr}

\DeclarePairedDelimiter{\ceil}{\lceil}{\rceil}

\usepackage{caption}
\usepackage{subcaption}

\usepackage{cite}

\ifCLASSINFOpdf
\else
\fi
\usepackage{array}
\usepackage{fixltx2e}
\usepackage{url}

\begin{document}
%
\title{Source localization using particle filtering on FPGA for robotic navigation with imprecise binary measurement
}
%
%
%

\author{Adithya Krishna, Andr\'e van Schaik,~\IEEEmembership{Fellow,~IEEE}, and Chetan Singh Thakur,~\IEEEmembership{Senior Member,~IEEE}
\thanks{A. Krishna and C. S. Thakur are with the Department of Electronic Systems Engineering, Indian Institute of Science, Bangalore - 560012, India (Email: \{adithyaik,csthakur\}@iisc.ac.in); A. van Schaik is with the International Centre for Neuromorphic Systems, The MARCS Institute, Western Sydney University, Australia. (Email: A.VanSchaik@westernsydney.edu.au);}}

\maketitle

\begin{abstract}

Particle filtering is a recursive Bayesian estimation technique that has gained popularity recently for tracking and localization applications. It uses Monte Carlo simulation and has proven to be a very reliable technique to model non-Gaussian and non-linear elements of physical systems. Particle filters outperform various other traditional filters like Kalman filters in non-Gaussian and non-linear settings due to their non-analytical and non-parametric nature. However, a significant drawback of particle filters is their computational complexity, which inhibits their use in real-time applications with conventional CPU or DSP based implementation schemes. This paper proposes a modification to the existing particle filter algorithm and presents a high-speed and dedicated hardware architecture. The architecture incorporates pipelining and parallelization in the design to reduce execution time considerably. The design is validated for a source localization problem wherein we estimate the position of a source in real-time using the particle filter algorithm implemented on hardware. The validation setup relies on an Unmanned Ground Vehicle (UGV) with a photodiode housing on top to sense and localize a light source. We have prototyped the design using Artix-7 field-programmable gate array (FPGA), and resource utilization for the proposed system is presented. Further, we show the execution time and estimation accuracy of the high-speed architecture and observe a significant reduction in computational time. Our implementation of particle filters on FPGA is scalable and modular, with a low execution time of about $5.62$ $\mu$s for processing $1024$ particles and can be deployed for real-time applications.

\end{abstract}


%
\IEEEpeerreviewmaketitle

\section{Introduction}
Emergency response operations such as disaster relief and military applications often require localization of a contaminant chemical or biological source in an unknown environment. Unmanned vehicles are gaining popularity in such applications in recent times due to reduced human involvement and ability to carry out the task remotely. These autonomous systems can eventually supplement human intervention in various safety-critical and hazardous missions. Nevertheless, conditions in which these missions are conducted vary drastically depending on the environmental factors, that result in the sensor receiving noise-corrupted measurements. This poses a significant challenge to the unmanned vehicles to navigate and locate a target in an unknown environment autonomously.
\par In our study, we demonstrate autonomous source localization using a UGV (robot) utilizing the Bearings-only tracking (BOT) model \cite{Farina} for light source localization with noise-corrupted input measurements as a proof-of-concept. The BOT model utilized here uses a particle filter algorithm \cite{Gordon} to increase the robustness to false detections, and noise-corrupted measurements. In recent times, there is growing popularity of particle filters (PFs) in signal processing and communication applications to solve various state estimation problems like tracking \cite{Tian}, localization, navigation \cite{Nicolas}, and fault diagnosis \cite{Zhang_c}. PFs have been applied to achieve filtering for models described using a dynamic state-space approach comprising a system model describing the state evolution and a measurement model describing the noisy measurements of the state\cite{Doucet}. In most real-time scenarios, these models are non-linear and non-Gaussian. Traditional filters like Kalman filters prove to be less reliable for such applications, and it has been shown that PFs outperform conventional filters in such scenarios \cite{Arulampalam}.
\par PFs are inherently Bayesian in nature, intending to construct a posterior density of the state (e.g., location of a target or source) from observed noisy measurements. In PFs, posterior of the state is represented by a set of weighted random samples known as particles. A weighted average of the samples gives the state estimate (location of the source). PFs use three major steps: Sampling, Importance, and Re-sampling for state estimation, thus deriving the name SIR filter. In the sampling step, particles from the prior distribution are drawn. The importance step is used to update the weights of particles based on input measurements. The re-sampling step prevents any weight degeneracy by discarding particles with lower weights and replicating particles having higher weights. Since PFs apply a recursive Bayesian calculation, all particles must be processed for sampling, importance, and re-sampling steps. Then, the process is repeated for the next input measurement, resulting in enormous computational complexity. Further, execution time of PFs is proportional to the number of particles, which inhibits the use of PFs in various real-time applications wherein a large number of particles need to be processed to obtain a good performance. Various implementation strategies (discussed below) have been suggested in the literature to alleviate this problem and make the PFs feasible in real-time applications.

\subsection{State-of-the-art}
 The first hardware prototype for PFs was proposed by Athalye et al. \cite{Athalye} by implementing a standard SIR filter on an FPGA. They provided a generic hardware framework for realizing SIR filters and implemented traditional PFs without parallelization on FPGA. As an extension to \cite{Athalye}, Bolic et al.\cite{Bolic} suggested a theoretical framework for parallelizing the re-sampling step by proposing distributed algorithms called Re-sampling with Proportional Allocation (RPA) and Re-sampling with Non-proportional Allocation (RNA) of particles to minimize execution time. The design complexity of RPA is significantly higher than that of RNA due to non-deterministic routing and complex routing protocol. Though the RNA solution is preferred over RPA for high-speed implementations with low design time, the RNA algorithm trades performance for speed improvement. Miao et al.  \cite{Miao1} proposed a parallel implementation scheme for PFs using multiple processing elements (PEs) to reduce execution time. However, the communication overhead between the PEs increases linearly with the number of PEs, making the architecture not scalable to process a large number of particles. Agrawal et al. \cite{Agrawal} proposed an FPGA implementation of a PF algorithm for object tracking in video. Ye and Zhang \cite{Ye} implemented an SIR filter on the Xilinx Virtex-5 FPGA for bearings-only tracking applications. Sileshi et al.  \cite{Sileshi,Sileshi1,Sileshi2} suggested two methods for implementation of PFs on an FPGA: the first method is a hardware/software co-design approach for implementing PFs using MicroBlaze soft-core processor, and the second approach is a full hardware design to reduce execution time. Velmurugan \cite{Velmurugan} proposed an FPGA implementation of a PF algorithm for tracking applications without any parallelization using the Xilinx system generator tool.
 \par Further, several real-time software-based implementation schemes have been proposed with an intent to reduce computational time. Hendeby et al. \cite{Hendeby} proposed the first Graphical Processing Unit (GPU) based PFs and showed that a GPU-based implementation outperforms a CPU implementation in terms of processing speed. Murray et al. \cite{Lawrence} provided an analysis of two alternative schemes for the re-sampling step based on Metropolis and Rejection samplers to reduce the overall execution time. They compared it with a standard Systematic resamplers \cite{Bolic_sr} over GPU and CPU platforms. Chitchian et al. \cite{Chitchian} devised an algorithm for implementing a distributed computation PF on GPU for fast real-time control applications. Zhang et al. \cite{Zhang} proposed an architecture for implementing PFs on a DSP efficiently for tracking applications in wireless networks. However, these software-based methods have their own drawbacks when it comes to hardware implementation owing to their high computational complexity. Therefore, it is essential to develop a high-speed and dedicated hardware design with the capacity to process a large number of particles in a specified time to meet the speed demands of real-time applications. This paper addresses this issue by proposing a high-speed architecture that is massively parallel and easily scalable to handle a large number of particles. The benefits of the proposed architecture are explained in the following subsection.


\subsection{Our contributions}
The contributions of this paper are on the algorithmic and hardware fronts: 

\subsubsection{Algorithmic Contribution}

We present an experimental framework and a measurement model to solve the source localization problem using the bearing-only tracking model in real-time. In our experiment, we employ a light source as the target/source to be localized and a UGV carrying an array of photodiodes to sense and localize the source. The photodiode measurements and the UGV position are processed to estimate the bearing of the light source relative to the UGV. Based on the bearing of the light source, we try to localize the source using the PF algorithm. Reflective objects and other stray light sources are also picked by the sensor (photodiodes), leading to false detections. In this study, we have successfully demonstrated that our PF system is robust to noise and can localize the source even when the environment is noisy. However, the PFs are computationally very expensive, and the execution time often becomes unrealistic using a traditional CPU based platform. The primary issue faced during the design of high-speed PF architecture is the parallelization of the re-sampling step. The re-sampling step is inherently not parallelizable as it needs the information of all particles. To overcome this problem, we propose a modification to the standard SIR filter (cf. Algorithm \ref{algo:sir_filter}) to make a parallel and high-speed implementation possible. The modified algorithm proposed (cf. Algorithm \ref{algo:sir_parallel}) uses a network of smaller filters termed as sub-filters, each doing the processing independently and concurrently. The processing of total $N$ particles is partitioned into $K$ sub-filters so that at most $N/K$ particles are processed within a sub-filter. This method reduces the overall computation time by a factor of $K$. The modified algorithm also introduces an additional particle routing step (cf. Algorithm  \ref{algo:sir_parallel}), which distributes the particles among the sub-filters and makes the parallel implementation of re-sampling possible. The particle routing step is integrated with the sampling step in the architecture proposed and does not require any additional time for computation. We also compare the estimation accuracy of the standard algorithm with the modified algorithm in Section \ref{sec:estimation_accuracy} and conclude that there is no significant difference in estimation error between the modified and the standard approach. Additionally, the modified algorithm achieves a very low execution time of about $5.62$  $\mu$s when implemented on FPGA, compared to $64$ ms on Intel Core i7-7700 CPU with eight cores clocking at $3.60$ GHz for processing $1024$ particles and outperforms other state-of-the-art FPGA implementation techniques.

\subsubsection{Hardware Contribution}

\par We implemented the proposed modified SIR algorithm on an FPGA, and key features of the architecture are:
\begin{itemize}[leftmargin=*]
\item Modularity: We divide the overall computation into multiple sub-filters, which process a fixed number of particles in parallel, and the processing of the particles is local to the sub-filter. This modular approach makes the design simple and adaptable as it allows us to customize the number of sub-filters in the design depending on the sampling rate of the input measurement and the amount of parallelism needed.
\item Scalability: Our architecture can be scaled easily to process a large number of particles without increasing the execution time, by using additional sub-filters. 
\item Design complexity: The proposed architecture relies on the exchange of particles between sub-filters. However, communication and design complexity increase proportionally with the number of sub-filters used in the design. In our architecture, we employ a simple ring topology to exchange particles between sub-filters to reduce complexity and design time. 
\item Memory utilization: The sampling step uses particles from the previous time instant to compute the particles of the current time instant. This requires the sampled and re-sampled particles to be stored in two separate memories. The straightforward implementation of the modified SIR algorithm needs $2 \times K$ memory elements each of depth $M$ for storing the sampled and re-sampled  particles for $K$ sub-filters. Here, $M$ is the number of particles in a sub-filter ($M=N/K$). However, applications involving non-linear models require a large number of particles \cite{LMiao}. This would make the total memory requirement $2 \times K $ significant for large $ K $ or $ M $. The proposed architecture reduces this memory requirement to $K$ memory elements each of depth $M$ using a dual-port ram, as explained in Section \ref{sec:sample}. Therefore, the proposed architecture reduces memory utilization and is more
energy-efficient due to reduced memory accesses. 
\item Real-time: Since all sub-filters operate in parallel, the execution time is significantly reduced compared to that of other traditional implementation schemes that use just one filter block \cite{Athalye}. Our implementation has a very low execution time of about $5.62$ $\mu$s (i.e., a sampling rate of $178$ kHz) for processing $1024$ particles and outperforms other state-of-the-art implementations, thus making it possible to be deployed in real-time scenarios. 
\item Flexibility: The proposed architecture is not limited to a single application, and the design can be easily modified by making slight changes to the architecture for other PF applications.

\end{itemize}

\par The architecture was successfully implemented on an Artix-7 FPGA. Our experimental results demonstrate the effectiveness of the proposed architecture for source localization using the PF algorithm with noise-corrupted input measurements.
\par The rest of the paper is organized as follows:  We provide the theory behind Bayesian filtering and PFs in Section \ref{sec:bayesframe} and \ref{sec:PF_bg}, respectively. The experimental setup for source localization using a Bearings-only tracking (BOT) model is presented in section \ref{sec:bot}. In this model, input to the filter is a time-varying angle (bearing) of
the source, and each input is processed by the PF algorithm implemented on hardware to estimate the source location. Further, in section \ref{sec:algo_mod}, we propose algorithmic modifications to the existing PF algorithm that make the high-speed implementation possible. The architecture for implementing PFs on hardware is provided in Section \ref{sec:arch_ov}. Evaluation of resource utilization on the Artix-7 FPGA, performance analysis in terms of execution time, estimation accuracy, and the experimental results are provided in Section \ref{sec:results}.

\section{Bayesian Framework}
\label{sec:bayesframe}
In a dynamic state space model, the evolution of the state sequence $x_{t}$ is defined by:
\begin{equation}
x_{t} = f_{t}(x_{t-1},w_{t})
\label{eqn:smdl}
\end{equation}
where, $f_{t}$ is a nonlinear function of the state $x_{t-1}$, and $w_{t}$ represents the process noise.
The objective is to recursively estimate the state $x_{t}$ based on a measurement defined by:
\begin{equation}
z_{t} = g_{t}(x_{t},v_{t})
\label{eqn:mmdl}
\end{equation}
where, $g_{t}$ is a nonlinear function describing the measurement model, and $z_{t}$ is the observation vector of the system corrupted by measurement noise $v_{t}$ at time instant $t$.
\par
From a Bayesian standpoint, the objective is to construct the posterior $p(x_{t}|z_{1:t})$ of the state $x_{t}$ from the measurement data $z_{1:t}$ up to time $t$. By definition, constructing the posterior is done in two stages namely, prediction and update.
\par The prediction stage involves using the system model (cf. Eq. \ref{eqn:smdl}) to obtain a prediction probability density function (PDF) of the state at time instant $t$, using the Chapman-Kolmogorov equation:
\begin{equation}
p(x_{t}|z_{1:t-1})=\sum _{x_{t-1}} p(x_{t}|x_{t-1})p(x_{t-1}|z_{1:t-1})
\label{eqn:ckeq}
\end{equation}  
where, the transition probability $p(x_{t}|x_{t-1})$ is defined by system model (cf. Eq. \ref{eqn:smdl}).
\par In the update stage, the measurement data $z_{t}$ at time step $t$ is used to update the PDF obtained from the prediction stage using Bayes rule, to construct the posterior:
\begin{equation}
p(x_{t}|z_{1:t})=\frac{p(z_{t}|x_{t})p(x_{t}|z_{1:t-1})}{\sum _{x_{t}}  p(z_{t}|x_{t})p(x_{t}|z_{1:t-1})}
\label{eqn:bayesr}
\end{equation}
\par where, $p(z_{t}|x_{t})$ is a likelihood function defined by the measurement model  (cf. Eq. \ref{eqn:mmdl}).
\par The process of prediction (cf. Eq. \ref{eqn:ckeq}) and update (cf. Eq. \ref{eqn:bayesr}) are done recursively for every new measurement $z_{t}$. Constructing the posterior based on Bayes rule is a conceptual solution and is analytically estimated using traditional Kalman filters. However, in a non-Gaussian and non-linear setting, the analytic solution is intractable, and approximation-based methods such as PFs are employed to find an approximate Bayesian solution. A detailed illustration of the Bayesian framework and its implementation for estimating the state of a system is provided by Thakur et al. \cite{Thakur}.

\section{Particle Filters Background }
\label{sec:PF_bg}
The key idea of PFs is to represent the required posterior density by a set of random samples termed as particles, with associated weights, and to compute the state estimate based on these particles and weights. 
The particles and their associated weights are denoted by  $\{x_{t}^{i},w_{t}^{i}\}_{i=1}^{N}$, where $N$ is the total number of particles. $x_{t}^{i}$ denotes the $i^{th}$ particle at time instant $t$. $w_{t}^{i}$ represents the weight corresponding to the particle $x_{t}^{i}$.
The most commonly used PF named the sampling, importance, and re-sampling filter (SIRF) is presented in Algorithm \ref{algo:sir_filter}.
 \begin{algorithm}[h]
\caption{: \,SIR Algorithm}
\begin{flushleft}
\textbf{Initialization:} Set the particle weights of previous time step to 1/N,  $\{w_{t-1}^{i}\}_{i=1}^{N} = 1/N$. \\
\textbf{Input:} Particles from previous time step $\{ x_{t-1}^{i} \}_{i=1}^{N}$ and measurement $z_{t}$.\\
\textbf{Output:} Particles of current time step  $\{\widehat{{x}}_{t}^{i}\}_{i=1}^{N} $. \\
\textbf{Method:} \\ 
\end{flushleft}
\begin{algorithmic}[1]
\\ \textbf{Sampling and Importance:}
\\ \hspace{0.3 cm} \text{for $i = 1$ to $N$ do}
\\ \hspace{0.75 cm} Sample $x_{t}^{i} \sim p(x_{t}|x_{t-1}^{i})$
\\ \hspace{0.75 cm} Calculate $w_{t}^{i} =  w_{t-1}^{i}p(z_{t}|x_{t}^{i})$
\\ \hspace{0.3 cm} \text{end for}
\\\textbf{Re-sampling:} Deduce the re-sampled particles $\{\widehat{{x}}_{t}^{i}\}_{i=1}^{N}$ from $\{x_{t}^{i},w_{t}^{i} \}_{i=1}^{N}$. 
\end{algorithmic}
\label{algo:sir_filter}
\end{algorithm}
\par In the sampling step, particles are drawn from the prior density $p(x_{t}|x_{t-1}^{i})$ to generate particles at time instant $t$. $p(x_{t}|x_{t-1}^{i})$ is deduced from (\ref{eqn:smdl}). Intuitively, it can be thought as propagating the particles from time step $t-1$ to $t$. The sampled set of particles at time instant $t$ is denoted by $\{{x}_{t}^{i}\}_{i=1}^{N}$. At time instant $0$,  particles are initialized with prior distribution to start the iteration. These particles are then propagated in time successively.
\par The importance step assigns weights to every particle $x_{t}^{i}$ based on the measurement $z_{t}$. By definition, the weights are given by:
\begin{equation}
w_{t}^{i} = w_{t-1}^{i}p(z_{t}|x_{t}^{i})
\label{eqn:wgteq1}
\end{equation}
However, weights of the previous time step are initialized to $1/N$ i.e $w_{t-1}^{i}=1/N$. Thus, we have:
\begin{equation}
w_{t}^{i} \propto p(z_{t}|x_{t}^{i})
\label{eqn:wgteq2}
\end{equation}
\par The re-sampling step is used to deal with the degeneracy problem in PFs. In the re-sampling step, particles with lower weights are eliminated, and particles with higher weights are replicated to compensate for the discarded particles depending on the weight $w_{t}^{i}$ associated with the particle $x_{t}^{i}$. The re-sampled set of particles is denoted by $\{\widehat{{x}}_{t}^{i}\}_{i=1}^{N}$.


\section{Bearings-Only Tracking Model}
\label{sec:bot}
This section briefly describes an experimental setup and a measurement model to solve the Bearings-Only Tracking (BOT) problem using the PF algorithm. As a real-world application, we consider a source localization problem wherein we try to estimate the source position based on the received sensor measurements in a noisy environment.

\subsection{Overview of the experimental setup} 
\begin{figure*}[h]
\centering
 \centering
  \begin{subfigure}[b]{0.52\textwidth}
    \includegraphics[width=\textwidth]{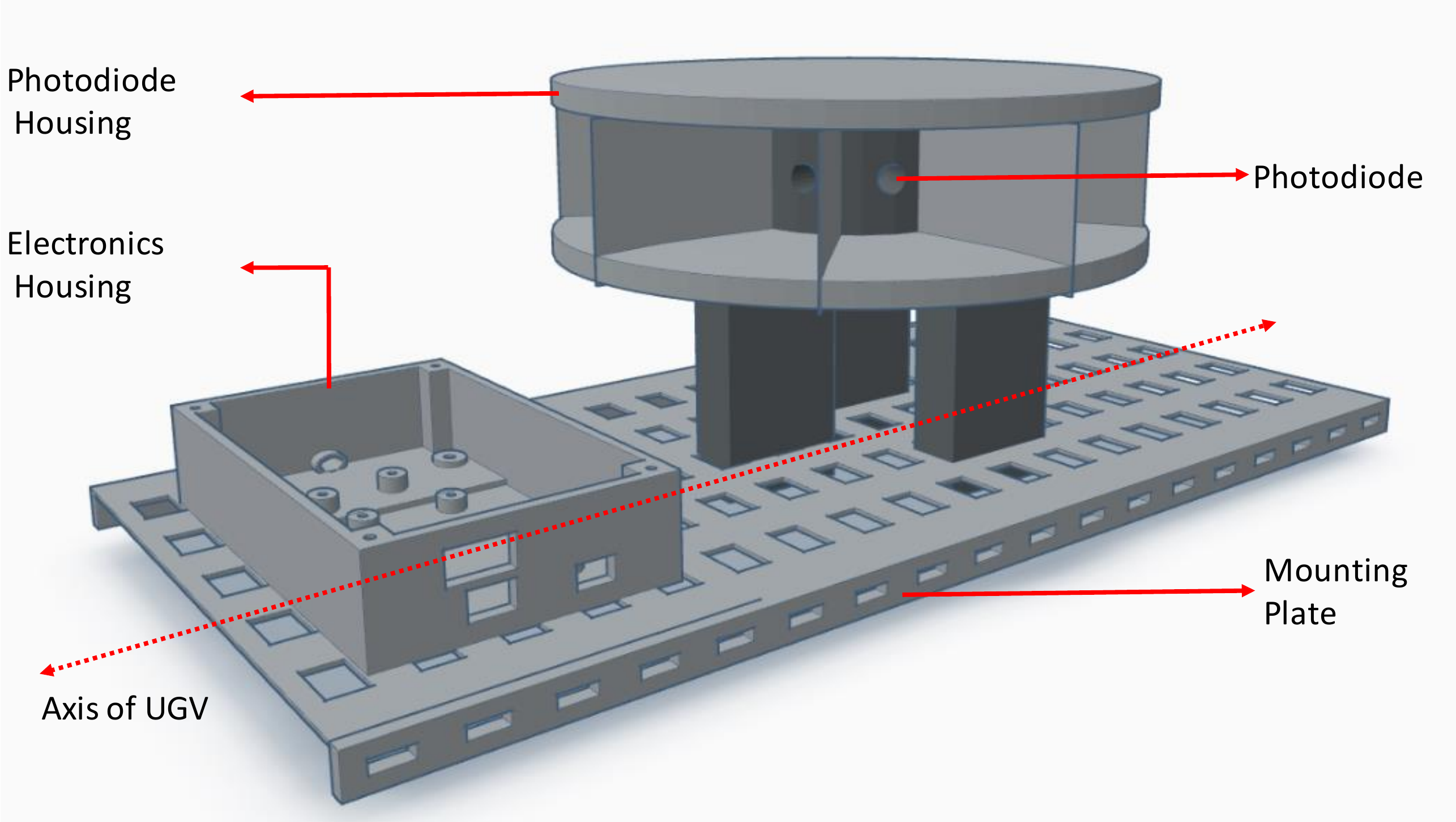}
    \caption{ }
  \end{subfigure}
  \begin{subfigure}[b]{0.27\textwidth}
    \includegraphics[width=\textwidth]{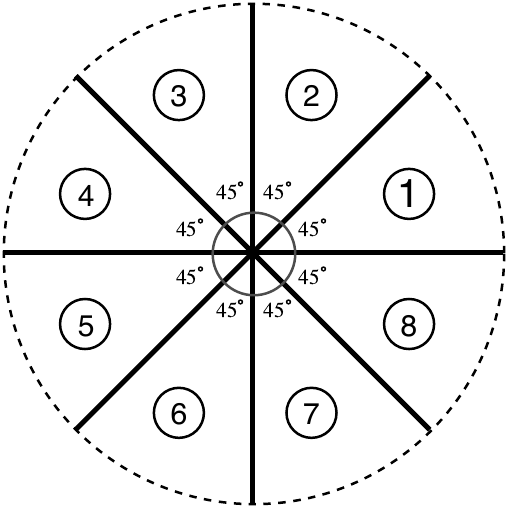}
    \caption{}
  \end{subfigure}
   \caption{UGV Design. (a) Schematic of the UGV with a photodiode housing mounted on top. (b) The region around the UGV is divided into $8$ sectors with $45\degree$ angular separation. }
  \label{fig:UGV_design}
\end{figure*}

In our experiment, an omnidirectional light source serves as a source to be localized. A photodiode housing mounted on top of a UGV (cf. Fig. \ref{fig:UGV_design}(a)) constitutes a sensor to measure the relative intensity of light in a horizontal plane. The space around the UGV is divided into $8$ sectors with $45\degree$ angular separation, as shown in Fig. \ref{fig:UGV_design}(b), and an array of $8$ photodiodes are placed inside the circular housing to sense the light source in all directions. The housing confines the angle of exposure of the photodiode to $45\degree$. Depending on the light incident on each photodiode, we consider the output of the photodiode to be either $0$ or $1$. 

The PF algorithm applied to the BOT model requires dynamic motion between the sensor and source \cite{Palmer}. We consider a stationary source, and a moving sensor mounted on a UGV, in our experimental setup. The UGV is made to traverse in the direction of the source and eventually converges to the source. Reflective sources and other stray light sources are potential sources of noise picked up by the sensor, producing false detections. A target-originated measurement, along with noise, is sensed by the photodiodes and processed in addition to the UGV position data, to estimate a bearing of the light source relative to the UGV. Based on the bearing of the light source, we try to estimate its position using the PF algorithm.

\subsection{Measurement model} 
\label{sec:measurement_model}
The position of UGV $(x_{t}^{UGV})$ at time instant $t$ is defined by the Cartesian co-ordinate system:
\begin{equation*}
x_{t}^{UGV} = [X_{t}^{UGV},Y_{t}^{UGV}]
\label{eqn:robcorn}
\end{equation*}
The orientation of the longitudinal axis of UGV is represented by $\phi_{t}^{UGV}$ which gives its true bearing.
\par The source is considered to be stationary, and its co-ordinates in the 2-dimensional setting is given by:
\begin{equation}
x_{t} = [X_{t},Y_{t}]
\label{eqn:tarcordn}
\end{equation}
At time instant $t$, a set of $8$ photodiode measurements are captured $z_{t}= \{ z_{t}^{1},z_{t}^{2} \cdots z_{t}^{8} \}$, which comprise of the target-associated measurement and clutter noise. Then, based on the measurement model \eqref{eqn:mmdl}, the source-associated measurement can be modelled as:
\begin{equation}
z_{t} = g(x_{t})+v_{t}
\label{eqn:mmdlBOT}
\end{equation}
Since the measurement gives the bearing information of the source, we have:
\begin{equation}
g(x_{t}) = tan^{-1}\left(\frac{Y_{t} - Y_{t}^{UGV}}{X_{t} - X_{t}^{UGV}} \right) 
\label{eqn:likelihoodfn}
\end{equation}
The four-quadrant inverse tangent function evaluated from $[0, 2\pi)$ gives the true bearing of the source.
\par The relevant probabilities needed to model the sensor imperfections and clutter noise are as follows:
\begin{enumerate}[label=(\roman*)]
\item The probability of clutter noise $(n_{t})$ produced by a stray or reflective light source is: $p(n_{t})=\beta$. 
\item The probability of the $j^{th}$ photodiode output being $1$ i.e., $(z_{t}^{j}=1)$ either due to light source or clutter noise is: $p(z_{t}^{j}|x_{t},n_{t})=\alpha$. 
\item If there is a light source in the sector $j$, then $j^{th}$ photodiode output will be $1$ with a probability of $\alpha$ irrespective of noise. The likelihood of photodiode output being $1$ or $0$ in the presence of the source is:
\begin{equation}
\label{eqn:l_tar}
  p(z_{t}^{j}|x_{t})=\begin{cases}
    \alpha, & \text{for $z_{t}^{j}=1 $}.\\
     1-\alpha, & \text{for $z_{t}^{j}=0$}.
  \end{cases}
\end{equation}

\item If there is no source in sector $j$, then there is a noise source with probability $\beta$. The likelihood of photodiode output being $1$ or $0$ in the absence of source is:
\begin{equation}
\label{eqn:l_notar}
  p(z_{t}^{j}|\widetilde{x_{t}})=\begin{cases}
    \alpha\beta, & \text{for $z_{t}^{j}=1 $}.\\
     1-\alpha \beta, & \text{for $z_{t}^{j}=0$}.
  \end{cases}
\end{equation}

\end{enumerate}
These two likelihoods are used in our system to model the sensor imperfections and noise, and even with high noise probability $\beta$, the PF algorithm is robust enough to localize the source.

\section{Algorithmic modification of SIRF for realizing high-speed architecture}
\label{sec:algo_mod}
In this section, we suggest modifications to the standard SIR algorithm to make it parallelizable. The key idea of the high-speed architecture is to utilize multiple parallel filters, termed as sub-filters, working simultaneously, and performing sampling, importance and re-sampling operations independently on particles. The architecture utilizes $K$ sub-filters in parallel to process a total of $N$ particles. Thus, the number of particles processed within each sub-filter is $M=N/K$. In this way, the number of particles processed within each sub-filter is reduced by a factor $K$, compared to conventional filters.
\begin{algorithm}[h]
\caption{: \,High-level description of each sub-filter $k$ performing SIR and particle routing operations.}
\begin{flushleft}
\textbf{Initialization:} Set the particle weights of previous time step to 1/M,  $\{w_{t-1}^{(k,i)}\}_{i=1}^{M} = 1/M$. \\
\textbf{Input:} Particles from previous time step $\{ x_{t-1}^{(k,i)}\}_{i=1}^{M}$  and measurement $z_{t}$\\
\textbf{Output:} Particles of current time step  $\{\widehat{{x}}_{t}^{(k,i)} \}_{i=1}^{M} $\\
\textbf{Method:} 
\end{flushleft}
\begin{algorithmic}[1]
\\\textbf{Particle Routing:} Exchange Q particles with neighbouring sub-filters. 

\\ \hspace{0.3 cm} $\{ x_{t-1}^{(k,q)} \}_{q=1}^{Q} \leftarrow  \{ x_{t-1}^{(k-1,q)} \}_{q=1}^{Q}$ for $k=2,\cdots K$, and \\ \hspace{0.25 cm}
$\{ x_{t-1}^{(k,q)} \}_{q=1}^{Q} \leftarrow \{ x_{t-1}^{(K,q)} \}_{q=1}^{Q}$ for k=1.
\\ \textbf{Sampling and Importance:}
\\ \hspace{0.3 cm} \text{for $i = 1$ to $M$ do}
\\ \hspace{0.75 cm} Sample $x_{t}^{(k,i)} \sim p(x_{t}|x_{t-1}^{(k,i)})$
\\ \hspace{0.75 cm}  Calculate $w_{t}^{(k,i)} = w_{t-1}^{(k,i)}p(z_{t}|x_{t}^{(k,i)})$
\\ \hspace{0.3 cm} \text{end for}
\\\textbf{Re-sampling:}  Compute the re-sampled particles $\{\widehat{{x}}_{t}^{(k,i)} \}_{i=1}^{M}$ from $\{x_{t}^{(k,i)},w_{t}^{(k,i)} \}_{i=1}^{M}$.

\end{algorithmic}
\label{algo:sir_parallel}
\end{algorithm}
 \par The sampling and importance steps are inherently parallelizable since there is no data dependency for particle generation and weight calculation. However, the re-sampling step cannot be parallelized as it needs to have the information of all particles. This creates a major bottleneck in the parallel implementation scheme. Thus, in addition to the SIR stage, we introduce a particle routing step, as shown in Algorithm \ref{algo:sir_parallel}, to route particles between sub-filters. The particle routing step enables the distribution of particles among sub-filters, and the re-sampling step can be effectively parallelized.
\par 
The particles and their associated weights in sub-filter $k$ at time step $t$ are represented by $\{ x_{t}^{(k,i)},w_{t}^{(k,i)} \}_{i=1}^{M}$, for $k=1,\cdots K$. The particle $x_{t}^{(k,i)}$ represents the position in the Cartesian co-ordinate system.

\section{Architecture Overview}
\label{sec:arch_ov}

In this section, we present a high-speed architecture for PFs, based on the modified SIRF algorithm presented in Section \ref{sec:algo_mod}.  
\begin{figure*}[h]
\centering
\includegraphics[width=16cm]{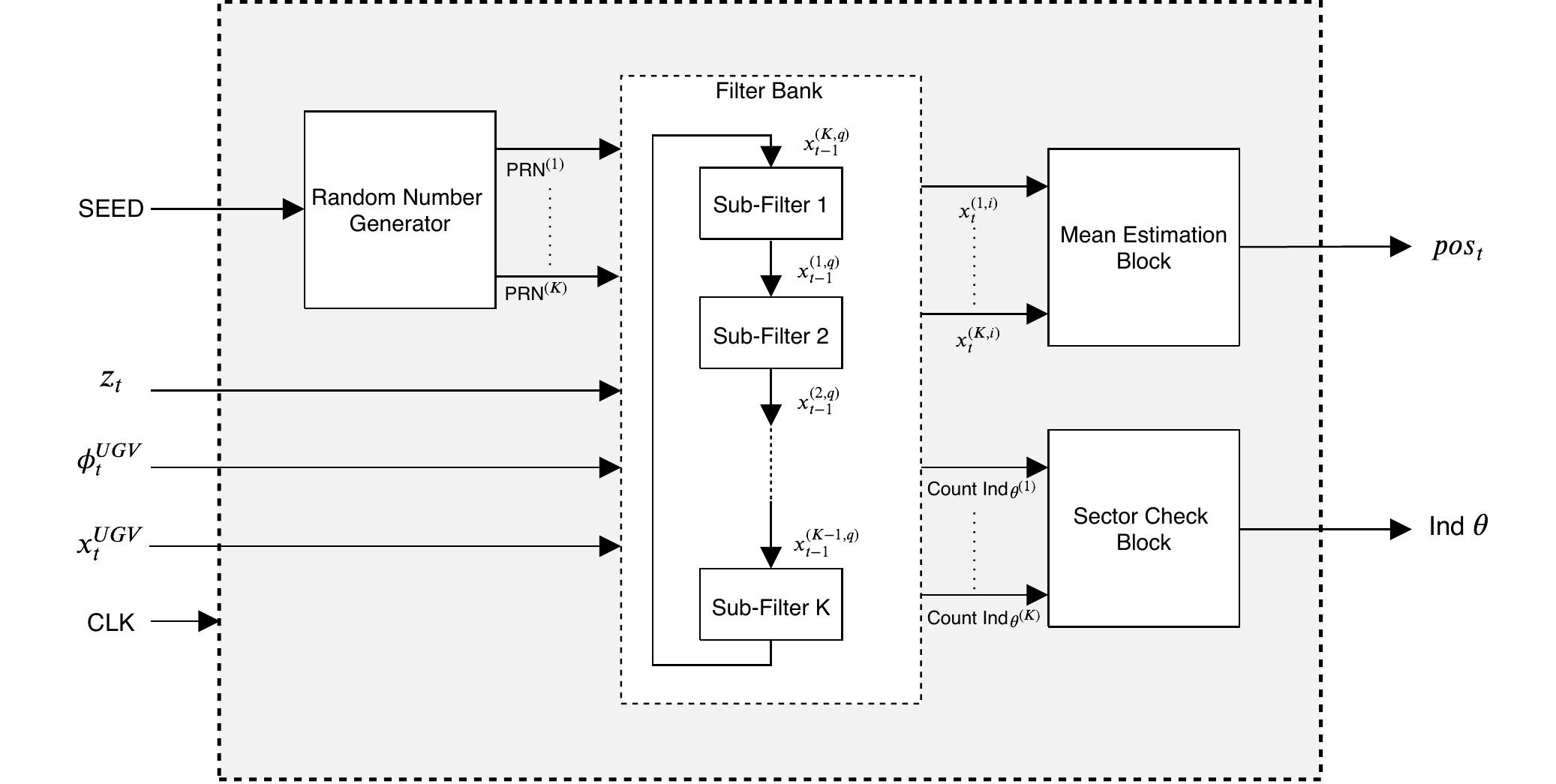}
\centering\caption{Top-level architecture of the realized particle filter algorithm.}
\label{fig:top_lvl_arch}
\end{figure*}
\par 
The top-level architecture shown in Fig. \ref{fig:top_lvl_arch} utilizes a filter bank consisting of $K$ sub-filters working in parallel. Sampling, importance, and re-sampling operations are carried out within a sub-filter. In addition to the SIR step, a fixed number of particles are routed between sub-filters after the completion of every iteration as part of a particle routing operation. The sub-filters are connected based on ring-topology inside the filter bank.  
$M$ particles are time-multiplexed and processed within each sub-filter, and $Q=M/2$ particles are exchanged with neighbouring sub-filters. Since the number of particles exchanged and the routing topology are fixed, the proposed architecture has very low design complexity. The design can be easily scaled up to process a large number of particles $(N)$ by replicating sub-filters. The binary measurements of the eight photodiodes $(z_{t})$ are fed as an input to the filter bank along with the true bearing  $(\phi_{t}^{UGV})$ and the position of the UGV $(x_{t}^{UGV})$. Random number generation needed for the sampling and re-sampling steps is provided by a random number generator block. We use a parallel multiple output LFSR architecture presented by Milovanovi\'{c} et al.  \cite{Milovanovic} for random number generation. As our internal variables are $16$ bits in size, a $16$ bit LFSR is used. Further, a detailed description of the sub-filter architecture is provided in Section \ref{sec:sub-filter}. The sector check block, described in Section \ref{sec:qc}, computes the particle population in each of the eight sectors and outputs a sector index that has the maximum particle population. This information is used by the UGV to traverse in the direction of the source. The mean computational block used to calculate the global mean of all $N$ particles from $K$ sub-filters to estimate the source location $(pos_{t})$, is explained in Section \ref{sec:meb}.

\begin {comment}

\subsection{Random Number Generator}
\label{sec:plfsr}
The traditional linear feedback shift register (LFSR) generates a single random number at particular time instant (say at each clock cycle). However, in the proposed architecture we utilize $K$ sub-filters in parallel thus requiring $K$ random numbers at each clock cycle. This can be achieved by utilizing $K$ LFSRs working in parallel with different seed or feedback polynomial, however this utilizes large number of registers (flip-flops) for higher value of $K$ and thus consuming more area. In order to overcome this problem, we use parallel multiple output LFSR architecture which generates multiple random numbers at every clock cycle utilizing a single LFSR as shown in the Fig. \ref{fig:plfsr}. The working principle of the architecture is similar to the conventional LFSRs, initially the contents of the LFSR is loaded to the EXOR network. The EXOR network is combinatorial logic which generates $K$ feedback terms in parallel. The architecture and the detailed explanation of the EXOR network is provided by Milovanovi\'{c} et al.  \cite{Milovanovic}.  The feedback points of the shift register known as taps is chosen based on the polynomial which generates maximum length sequence.  The feedback polynomial for generating maximal sequence is given by Alfke \cite{Alfke}. At the beginning of the next clock cycle contents of the LFSR is shifted by $K$ positions to the right and leftmost $K$ cells in the LFSR is loaded with the $K$ feedback terms. The LFSR contents and the feedback array consisting of feedback terms is loaded on to the output register at every clock cycle to produce $K$ parallel random numbers each of width $L$. 

\begin{figure*}[!t]
\centering
\includegraphics[width=14cm]{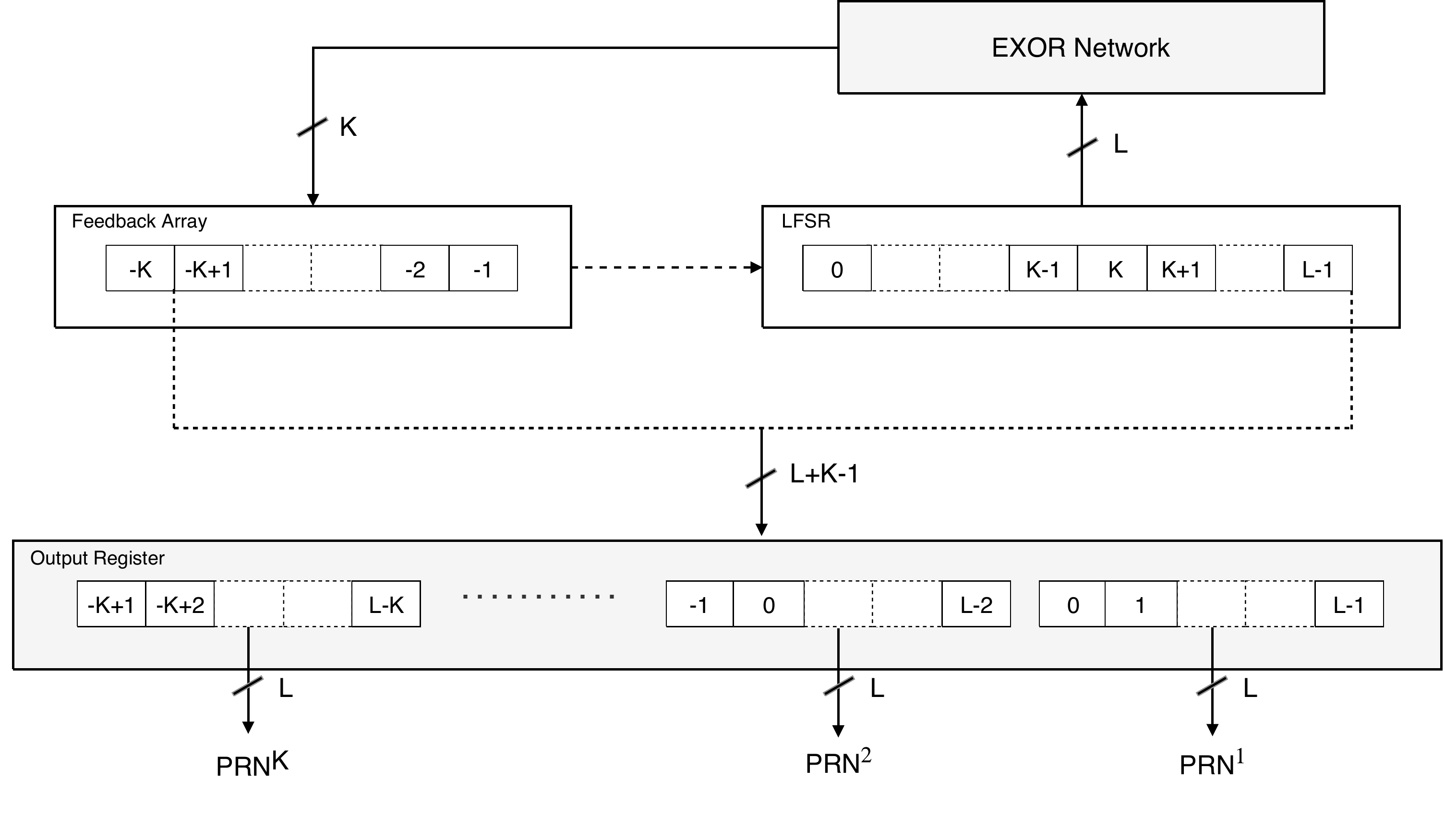}
\centering\caption{Random number generator architecture which generates multiple parallel random numbers.}
\label{fig:plfsr}
\end{figure*}
\end{comment}

\subsection{Sub-filter architecture}
\label{sec:sub-filter}
\begin{figure*}[h]
\centering
\includegraphics[width=13cm]{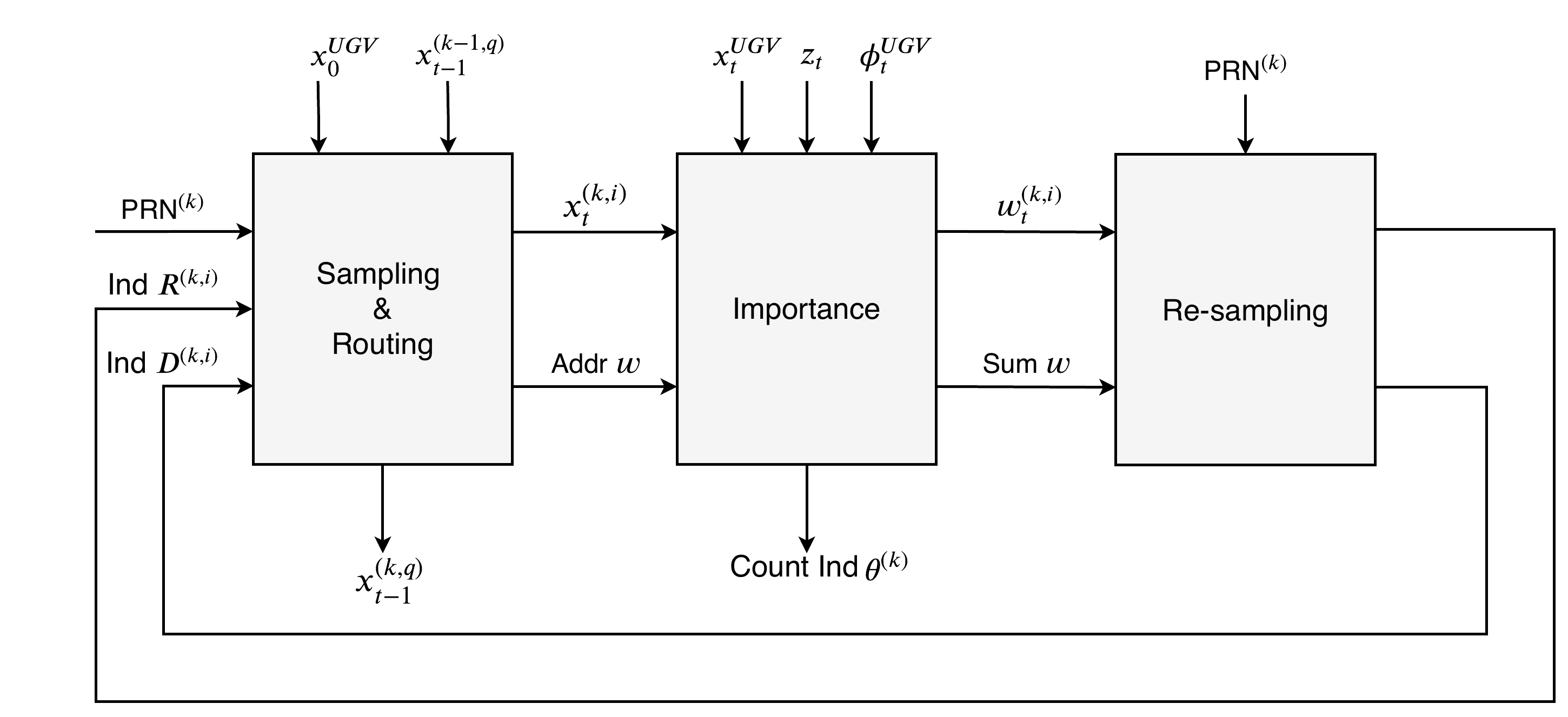}
\centering\caption{Sub-filter architecture.}
\label{fig:sub-filter}
\end{figure*}

The sub-filter is the main computational block responsible for particle generation, processing, and filtering. It consists of three main sub-modules, namely, sampling, importance and re-sampling, as shown in Fig. \ref{fig:sub-filter}. The sampling and importance blocks are pipelined in operation. The re-sampling step cannot be pipelined with the former steps as it requires weight information of all particles. Thus, it is started after the completion of the importance step. Since sampling and importance stages are pipelined, together they take $M$ clock cycles to iterate for $M$ particles, as shown in Algorithm \ref{algo:sir_parallel} from line $5$ to line $8$. The particle routing between the sub-filters is done along with the sampling step and does not require any additional cycles. The re-sampling step takes $3M$ clock cycles, as discussed in Section \ref{sec:resampling}.

\subsubsection{Sampling and routing}
\label{sec:sample}
The sampling step involves generating new sampled particles $\{ x_{t}^{(k,i)} \}_{i=1}^{M}$ by propagating the re-sampled particles $\{\widehat{{x}}_{t-1}^{(k,i)}\}_{i=1}^{M}$  from the previous time step using the dynamic state space model:
\begin{equation}
x_{t}^{(k,i)} \sim p(x_{t}|\widehat{x}_{t-1}^{(k,i)})
\end{equation}

\begin{figure*}[!t]
\centering
\includegraphics[width=16cm]{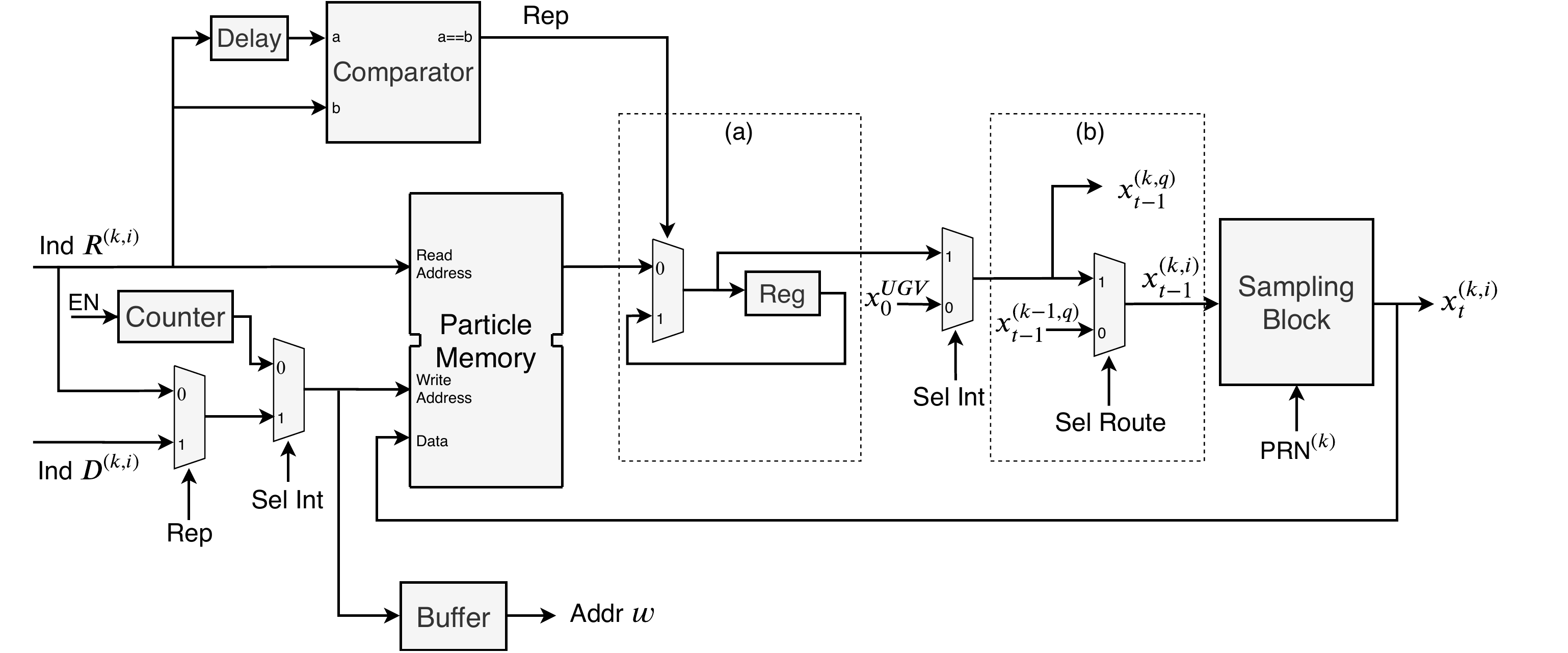}
\centering\caption{Sampling and routing unit architecture.}
\label{fig:sampling}
\end{figure*}

Conventionally, particles $\{ x_{t}^{(k,i)} \}_{i=1}^{M}$ are used to generate the weights $\{w_{t}^{(k,i)}\}_{i=1}^{M}$ in the importance unit, and using these weights we determine the re-sampled particles  $\{\widehat{{x}}_{t}^{(k,i)}\}_{i=1}^{M}$. Further,  $\{\widehat{{x}}_{t}^{(k,i)}\}_{i=1}^{M}$ is utilized to obtain particles $\{ x_{t+1}^{(k,i)} \}_{i=1}^{M}$ of the next time step.
Thus, with the straightforward approach, we would need two memories each of depth $M$ to store $\{ x_{t}^{(k,i)} \}_{i=1}^{M}$ and  $\{\widehat{{x}}_{t}^{(k,i)}\}_{i=1}^{M}$ within a sub-filter. Similarly, for $K$ sub-filters we would require $2\times K$ memory elements, each of depth $M$. This increases memory usage for higher $K$ or $M$. In this paper, we propose a novel scheme to store the particles using a single dual-port memory instead of two memory blocks, which brings down the total memory requirement for storing particles to $K$ memory elements, each of depth $M$. 
\par In this scheme, since the re-sampled particles are actually the subset of sampled particles $($i.e.,$ \{\widehat{{x}}_{t}^{(k,i)}\}_{i=1}^{M} \subset \{x_{t}^{(k,i)} \}_{i=1}^{M})$ instead of storing $\{\widehat{{x}}_{t}^{(k,i)}\}_{i=1}^{M}$ in a different memory, we can use the same memory as $\{x_{t}^{(k,i)} \}_{i=1}^{M}$ and use suitable pointers or indices to read $\{\widehat{{x}}_{t}^{(k,i)}\}_{i=1}^{M}$. 
\par The re-sampling unit in our case is modified such that instead of returning re-sampled particles $\widehat{{x}}_{t-1}^{(k,i)}$, it returns the indices of replicated (Ind $R^{(k,i)}$) and discarded (Ind $D^{(k,i)}$) particles (cf. Fig. \ref{fig:sub-filter}). Ind $R^{(k,i)}$ is used as a read address of the dual-port particle memory shown in Fig. \ref{fig:sampling} to point to the re-sampled particles $\widehat{{x}}_{t-1}^{(k,i)}$. The dual-port memory enables us to perform read and write operations simultaneously; however, this might result in data overwriting. For example, consider six particles, after re-sampling particle $2$ ($x_{t-1}^{(k,2)}$) is replicated four times; particle $5$ ($x_{t-1}^{(k,5)}$) is replicated two times and particles $1,3,4\,\&\,6$ are discarded. The re-sampling unit returns Ind $R = (2,2,2,2,5,5)$ and Ind $D = (1,3,4,6)$. The read sequence of the dual-port memory is $(2,2,2,2,5$ \& $5)$ and the write sequence is $(2,1,3,4,5$ \& $6) $. Initially, particle $2$ $(x_{t-1}^{(k,2)})$ is read from the dual-port memory and after propagation in the sampling block, the sampled particle $x_{t}^{(k,i)}$ is written back to the memory location $2$. Next, particle $2$ is read again from memory location $2$. However, this time the content of the location is changed and it no longer holds the original particle $x_{t-1}^{(k,2)}$, which causes an error while reading. In order to avoid this scenario, we introduce a sub-block (a) (cf. Fig. \ref{fig:sampling}), wherein when we read the particle from the memory for the first time, it is temporarily stored in a register. Hence, whenever there is a replication in Ind $R$ or read address, we read the particle from the register instead of memory. The Rep signal is generated by comparing Ind $R$ with its previous value and if both are same, Rep will be made high.
 \begin{algorithm}[h]
\caption{: \,Sampling block pseudocode}
\begin{flushleft}
\textbf{Input:} Particles from previous time step $ x_{t-1}^{(k,i)}=[X_{t-1}^{(k,i)} ,Y_{t-1}^{(k,i)}]$ and random number $PRN^{(k)}=[PRN_{x}^{(k)},PRN_{y}^{(k)}]$.\\
\textbf{Output:} Particles of current time step  $ x_{t}^{(k,i)}$. \\
\textbf{Method:} \\ 
\end{flushleft}
\begin{algorithmic}[1]
\\ \hspace{0.3 cm} \text{for $i = 1$ to $M$ do}
\\ \hspace{0.75 cm} $X_{t}^{(k,i)} = X_{t-1}^{(k,i)} + PRN_{x}^{(k)}*std $
\\ \hspace{0.75 cm} $Y_{t}^{(k,i)} = Y_{t-1}^{(k,i)} + PRN_{y}^{(k)}*std$  \algorithmiccomment{$std$ is the standard deviation.} 
\\ \hspace{0.75 cm}$x_{t}^{(k,i)} = [X_{t}^{(k,i)} ,Y_{t}^{(k,i)}]$
\\ \hspace{0.3 cm} \text{end for}
\end{algorithmic}
\label{algo:sampling_blk}
\end{algorithm}

\begin{figure*}[t]
\centering
\includegraphics[width=16cm]{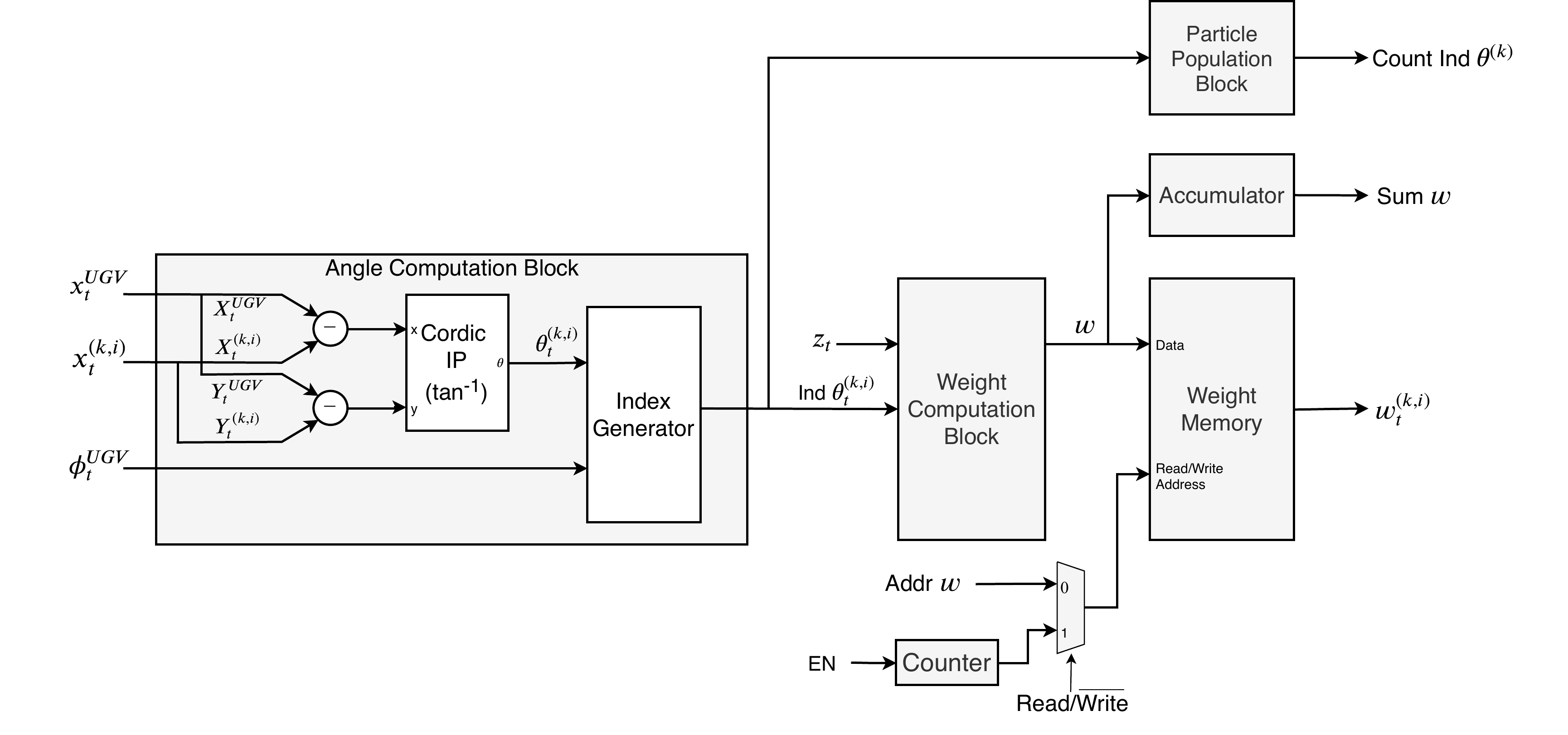}
\centering\caption{Importance unit architecture.}
\label{fig:imp}
\end{figure*}

\par Further, we introduce a sub-block (b) (cf. Fig. \ref{fig:sampling}), which is responsible for routing the particles between neighbouring sub-filters. Out of $M$ particles read from the particle memory of sub-filter $k$, the first $M/2$ particles, i.e., $\{x_{t-1}^{(k,q)}\}_{q=1}^{M/2}$ are sent to sub-filter $k+1$, and simultaneously the first $M/2$ particles, i.e., $\{x_{t-1}^{(k-1,q)}\}_{q=1}^{M/2}$, of sub-filter $k-1$ are read and fed to the sampling block of sub-filter $k$.
The sampling block propagates the particles from time step $t-1$ to time step $t$. The routed particles from sub-filter $k-1$ $\{x_{t-1}^{(k-1,q)}\}_{q=1}^{M/2}$, and last $M/2$ local particles $\{x_{t-1}^{(k,q)}\}_{q=M/2+1}^{M}$ read from particle memory of sub-filter $k$ are propagated by the sampling block and written back to the memory. The input to the sampling block are particles of time step $t-1$ $(x_{t-1}^{(k,i)})$ and the output are particles of current time step $t$ $(x_{t}^{(k,i)})$. The sampling block pseudocode is provided by Algorithm \ref{algo:sampling_blk}. The random number $PRN^{(K)}$ needed for random sampling of particles as shown in Algorithm \ref{algo:sampling_blk}, line $2$ and line $3$ is provided by a random number generator block (cf. Fig. \ref{fig:top_lvl_arch}). The $Sel\,Route$ signal is used to control the switching between the local and routed particles by making it low for the first $M/2$ cycles and then making it high for the next $M/2$ cycles. Further, at time instant $0$, we feed the UGV position $x_{0}^{UGV}$ as a prior to the sampling block to distribute the particles around the UGV. 
The $Sel\,Int$ control signal is made low in the first iteration, i.e., at time instant 0, and then made high for the subsequent iterations.

\subsubsection{Importance}
\label{sec:imp}

The importance unit computes the weights of the particles based on the photodiode measurements $z_{t}$ given by:

\begin{equation}
    w_{t}^{(k,i)} =  w_{t-1}^{(k,i)}p(z_{t}|x_{t}^{(k,i)})
\end{equation}

$w_{t-1}^{(k,i)}$ is initialized to 1/M. Estimation of  $p(z_{t}|x_{t}^{(k,i)})$ involves determining the angle of each particle $(\theta_{t}^{(k,i)}$), which is computed using an inverse tangent function based on the position of UGV $(x_{t}^{UGV})$ and position of the particle $(x_{t}^{(k,i)})$, as follows:
\begin{equation*}
    \theta_{t}^{(k,i)} = tan^{-1}\left(\frac{Y_{t}^{(k,i)}-Y_{t}^{UGV}}{X_{t}^{(k,i)}-X_{t}^{UGV}} \right) 
\end{equation*}
where, $X_{t}^{(k,i)}$ and $Y_{t}^{(k,i)}$ represents the co-ordinates of the particle $x_{t}^{(k,i)}$ in two-dimensional Cartesian co-ordinate system.

The inverse tangent function is implemented using a Cordic IP block provided by Xilinx \cite{Cordic}. The architecture of the importance unit is shown in Fig. \ref{fig:imp}. The index generator block estimates the angle of the particles with respect to the longitudinal axis of the UGV based on the bearing of the UGV  $(\phi_{t}^{UGV})$. In addition to this, the index generator block is used for determining the sector indices (Ind $\theta_{t}^{(k,i)}$) of the particles based on the angle information. The sector indices of the particle can be defined as follows:

\begin{equation*}
Ind\,\,\theta_{t}^{(k,i)} = \ceil{4/\pi*(\theta_{t}^{(k,i)}-\phi_{t}^{UGV})}
\end{equation*}

$z_{t}$ is $8$ bit wide data consisting of $8$ binary photodiode measurements  $\{ z_{t}^{1},z_{t}^{2} \cdots z_{t}^{8} \}$. Based on the measurement $z_{t}$ and the sector indices of the particles, weights are generated by the weight computation block. These weights are stored in the weight memory using the address provided by the sampling unit, to store weights in the same order as the sampled particles $x_{t}^{(k,i)}$. The sum of all the weights required by the re-sampling unit is obtained by an accumulator. The particle population block is used to estimate the number of particles present in each of the eight sectors, using the sector indices of the particles for a given sub-filter. The particle count in each of the eight sectors of sub-filter $k$ is concatenated and given as the output Count Ind $\theta^{(k)}$. For example, if sector $1$ has $15$ particles, sector $3$ has $14$ particles, and sector $5$ has $3$ particles, then Count Ind $\theta^{(k)} = \{15,0,14,0,3,0,0,0\}$.

\begin{figure*}[t]
\centering
\includegraphics[width=11cm]{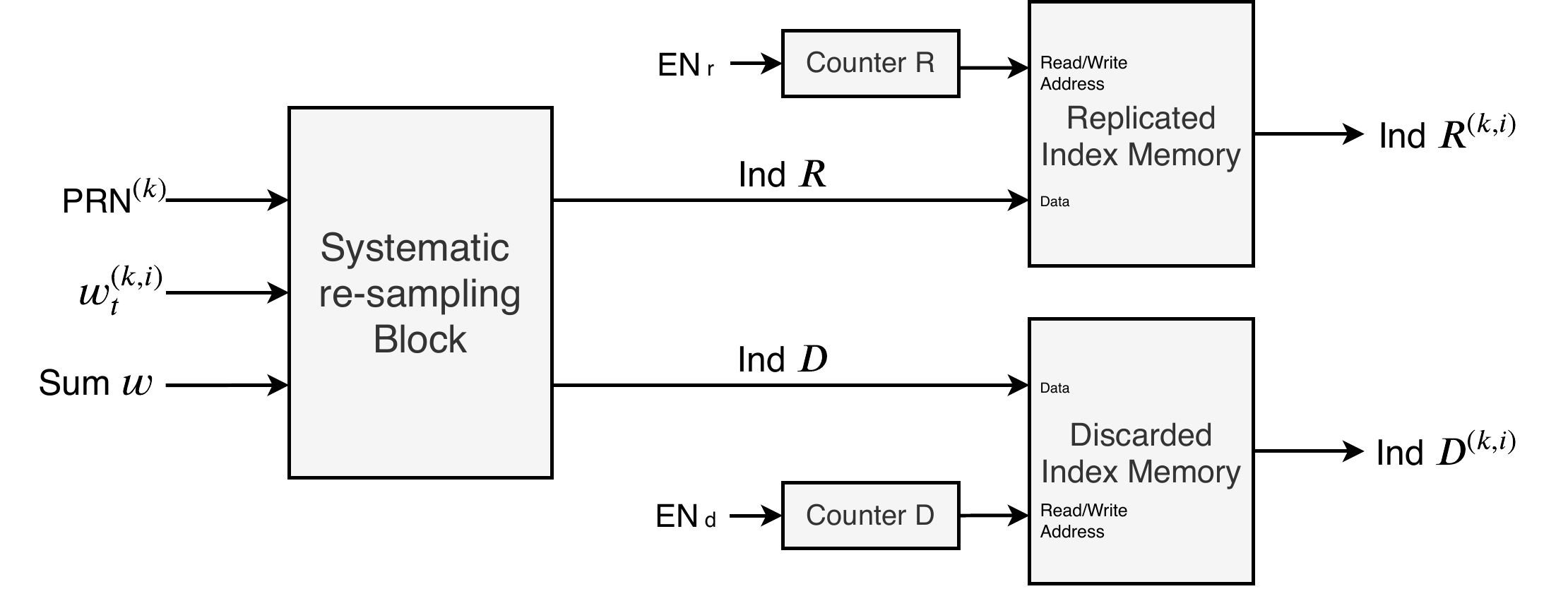}
\centering\caption{Re-sampling unit architecture.}
\label{fig:resample}
\end{figure*}

\subsubsection{Re-sampling}
\label{sec:resampling}

The re-sampling step replicates particles with higher weights and discards particles with lower weights. This is accomplished by utilizing a Systematic re-sampling algorithm shown in Algorithm \ref{algo:sr_algo}. A detailed description of the systematic re-sampling algorithm is provided in \cite{Athalye,Bolic_sr}. The weights and sum of all weights are obtained from the importance unit. The random number $(U_{0})$ needed to compute the parameter $U\_scale$ in line 2 of Algorithm \ref{algo:sr_algo} is provided by the Random number generator block shown in Fig. \ref{fig:top_lvl_arch}. The algorithm presented works with un-normalized weights, which will avoid M division operations on all particles to implement normalization. The division required to compute $A_w$ in line 1 of Algorithm \ref{algo:sr_algo} is implemented using the right shift operation. This approach consumes fewer resources and area on hardware. The replicated and discarded indices generated by the systematic re-sampling block are stored in their respective memories, as shown in Fig. \ref{fig:resample}. In the worst-case scenario, the inner loop of Algorithm \ref{algo:sr_algo} takes $2M$ cycles for execution in hardware as it involves fetching $M$ weights from weight memory and doing $M$ comparison operations. Further, line 13 and line 14 take $M$ cycles to obtain $M$ replicated indices. Thus, in total, the execution of the re-sampling step requires $2M+M=3M$ cycles.

\begin{algorithm}[h]
\caption{: \,Systematic Re-sampling.}
\begin{flushleft}       
\textbf{Input:} Un-normalized weights ($\{w_{t}^{(k,i)}\}_{i=1}^{M}$) of M particles, summation of all the weights in a sub-filter (Sum $w$) and the uniform random number ($U_{0}$) between $[0,1]$\\
\textbf{Output:}  Replicated index (Ind $R$) and Discarded index (Ind $D$).\\
\textbf{Method:} \\ 
\end{flushleft}
\begin{algorithmic}[1]
\\      
Compute $A_w = \dfrac{Sum\,w}{M}$\\
Initialize :   $U\_scale = U_{0} \times Aw$ \\ 
 \hspace{1.4 cm}  $s=0,p=0,m=0$  

\For{i=1 to M} 
 \While{$s<U\_scale$}
 \\ \hspace{1 cm}  $p = p+1$
  \\ \hspace{1 cm}  $s = s+w^{(k,p)}$
  \If{$s<U\_scale$}
   \\ \hspace{1.4 cm}  $m = m+1$
    \\ \hspace{1.54 cm}Ind $D^{(k,m)} = p$
  \EndIf
  \EndWhile
    \\ \hspace{1 cm}  $U\_scale = U\_scale+A_w$
     \\ \hspace{1 cm} Ind $R^{(k,i)} = p$
\EndFor
\end{algorithmic}
\label{algo:sr_algo}
\end{algorithm}

\subsection{Sector check block}
\label{sec:qc}
\begin{figure}[h]
\centering
\includegraphics[width=\columnwidth]{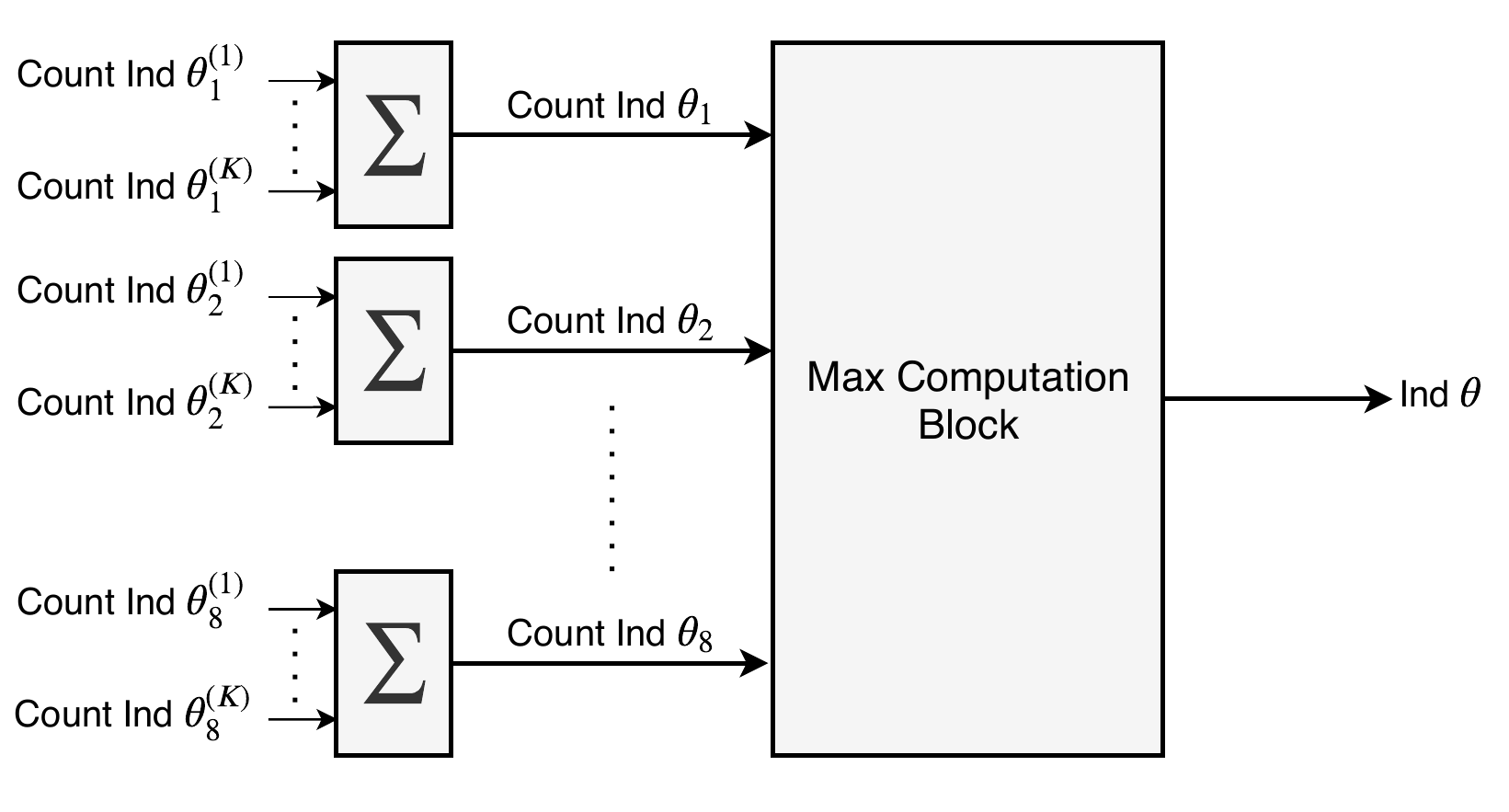}
\centering\caption{Sector check block architecture.}
\label{fig:qc}
\end{figure}

The direction/orientation of the UGV is decided by the population of particles in different sectors and is used to move towards the source. This is achieved by the sector check block, which estimates the particle population in each of the eight sectors and gives the sector index with maximum particle count.
The block diagram shown in Fig. \ref{fig:qc} utilizes eight parallel adders to count the number of particles in each sector. The particle count in a given sector of $K$ sub-filters is fed as an input to the adder. Count Ind $\theta_{n}^{(k)}$ in Fig. \ref{fig:qc} denotes the particle count in sector $n$ of sub-filter $k$. The output of an adder gives the total particle population in a particular sector. Furthermore, the sector index (Ind $\theta$) having maximum particle count is estimated using a max computation block. The UGV uses this information to traverse in the direction of the source. 

\subsection{Mean estimation block}
\label{sec:meb}
The mean of total $N$ particle positions is estimated using the mean estimation block. Particle positions from $K$ sub-filters are fed in parallel and accumulated over $M$ cycles to generate the sum, which is further divided by $N$, by right shifting $log_{2}(N)$  times to get the mean. In our implementation, we consider $N$ as a power of $2$. The mean gives an estimate of the position of the source $pos_{t}$.

\section{Results}
\label{sec:results}
In this section, we present resource utilization of the proposed design on an FPGA. We also evaluate the execution time of the proposed architecture as a function of number of sub-filters and inspect the estimation accuracy by scaling the number of particles. We then compare our design with state-of-the-art implementations. Furthermore, we present experimental results for the source localization problem implemented on an FPGA using the proposed architecture. 

\begin{table*}[h]
\caption{Resource utilization on Artix-7 FPGA.}
\label{tab:ru}

\centering
\begin{tabular}{M{0.14\textwidth}|M{0.14\textwidth}|M{0.14\textwidth}|M{0.14\textwidth}|M{0.14\textwidth}|M{0.14\textwidth}}
\hline \hline\\ [-1em]
\text{Sub-filters $K$} & \text{Occupied slices} & \text{Slice LUTs}  & \text{LUTRAM} & \text{Slice Registers} & \text{Block RAM} \\  \hline \hline \\ [-1em]
$1$ & $\shortstack{505 \\(1.5\%)}$ & $ \shortstack{1,437\\ (1.07\%)}$ & $\shortstack{51\\(0.11\%)}$ & $\shortstack{1,735\\(0.65\%)}$ & $\shortstack{2\\(0.55\%)}$   \\ \hline \\ [-1em]
 $2$ & $\shortstack{942\\(2.8\%)}$ & $\shortstack{2,847\\ (2.13\%)}$ & $\shortstack{102\\(0.22\%)}$ & $\shortstack{3,259\\(1.22\%)}$ & $\shortstack{4\\(1.10\%)}$ \\ \hline \\ [-1em]
 $4$ & $\shortstack{1,710\\(5.1\%)}$ & $\shortstack{5,494\\ (4.11\%)}$ & $\shortstack{204\\(0.44\%)}$ & $\shortstack{6,297\\(2.35\%)}$ & $\shortstack{8\\(2.19\%)}$\\ \hline \\ [-1em]
 $8$ & $\shortstack{3,465\\(10.3\%)}$ & $\shortstack{10,973\\ (8.20\%)}$ & $\shortstack{408\\(0.88\%)}$ & $\shortstack{12,356\\(4.62\%)}$ & $\shortstack{16\\(4.38\%)}$ \\ \hline \\ [-1em]
 $16$ & $\shortstack{6,834\\(20.3\%)}$ & $\shortstack{21,885\\ (16.36\%)}$ & $\shortstack{816\\(1.77\%)}$ & $\shortstack{24,460\\(9.14\%)}$ & $\shortstack{32\\(8.77\%)}$  \\ \hline \\ [-1em]
 $32$ & $\shortstack{14,513\\(43.1\%)}$ & $\shortstack{43,804\\ (32.74\%)}$ & $\shortstack{1,632\\ (3.53\%)}$ & $\shortstack{48,645\\(18.18\%)}$ & $\shortstack{64\\(17.53\%)}$\\ \hline \\ [-1em]
 $64$ & $\shortstack{29,290\\(87\%)}$ & $\shortstack{86,101\\ (64.35\%)}$ & $\shortstack{3,264\\(7.06\%)}$ & $\shortstack{95,570\\(35.71\%)}$ & $\shortstack{128\\ (35.07\%)}$ \\ \hline \hline
\end{tabular}
\label{tab:SQNR}
\end{table*}
   
\subsection{Resource utilization}
The architecture presented was implemented on an Artix-7 FPGA. Resource utilization of the implemented design for different number of sub-filters is summarized in Table \ref{tab:ru}. The number of particles per sub-filter $(M)$ was fixed to $32$ for synthesizing the design. All memory modules shown in the architecture for storing particles, weights, replicated, and discarded indices are mapped into embedded 18kb block random access memory (BRAM) available on the FPGA, using a Block Memory Generator (BMG) IP \cite{bmg} provided by Xilinx. The Block RAM column in Table \ref{tab:ru} indicates the number of 18kb BRAM blocks needed on the FPGA. 
It can be seen that resource utilization increases proportionally with the number of sub-filters. For $64$ sub-filters, $64\%$ of the slice LUTs (lookup tables) are used, and a maximum of approximately $90$ sub-filters can fit onto a single Artix-7 (xc7a200tfbg484-1) FPGA platform.


\subsection{Execution time}
\par The proposed design utilizes $K$ parallel sub-filters, thus bringing down the number of particles processed within a sub-filter to $N/K$. Since sampling and importance blocks are pipelined, these steps take $N/K+\tau_{s}+\tau_{i}$ clock cycles and the re-sampling step takes $3N/K+\tau_{r}$ cycles to process $N/K$ particles, where $\tau_{s}$ , $\tau_{i}$ and $\tau_{r}$ represent the start-up latency of the sampling, importance and re-sampling units, respectively. Since all the $K$ sub-filters are parallelized, the time taken to process a total of $N$ particles for the SIR operation is:
\begin{equation*}
T_{SIR} = (4N/K+\tau)T_{clk}   
\end{equation*}
where, $\tau=\tau_{s}+\tau_{i}+\tau_{r}$ and $T_{clk}$ is the clock period of the design.
\par
\begin{figure*}[h]
\centering
\includegraphics[width=6.6cm]{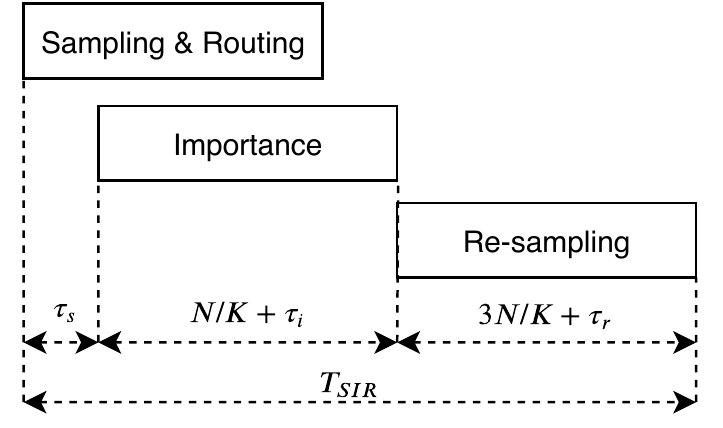}
\centering\caption{Timing diagram for SIR operations of the proposed design.}
\label{fig:td}
\end{figure*}

 Fig. \ref{fig:td} gives the timing diagram for completion of SIR operations using the proposed architecture for $N$ particles, for a single iteration. Furthermore, since particle routing is incorporated within the sampling step, the transfer of particles between the sub-filters does not take any additional cycles. This makes the design scalable for a large number of sub-filters, as the routing operation requires no extra time.

\par
\begin{figure*}[h]
  \centering

  \begin{subfigure}[b]{0.48\textwidth}
    \includegraphics[width=\textwidth]{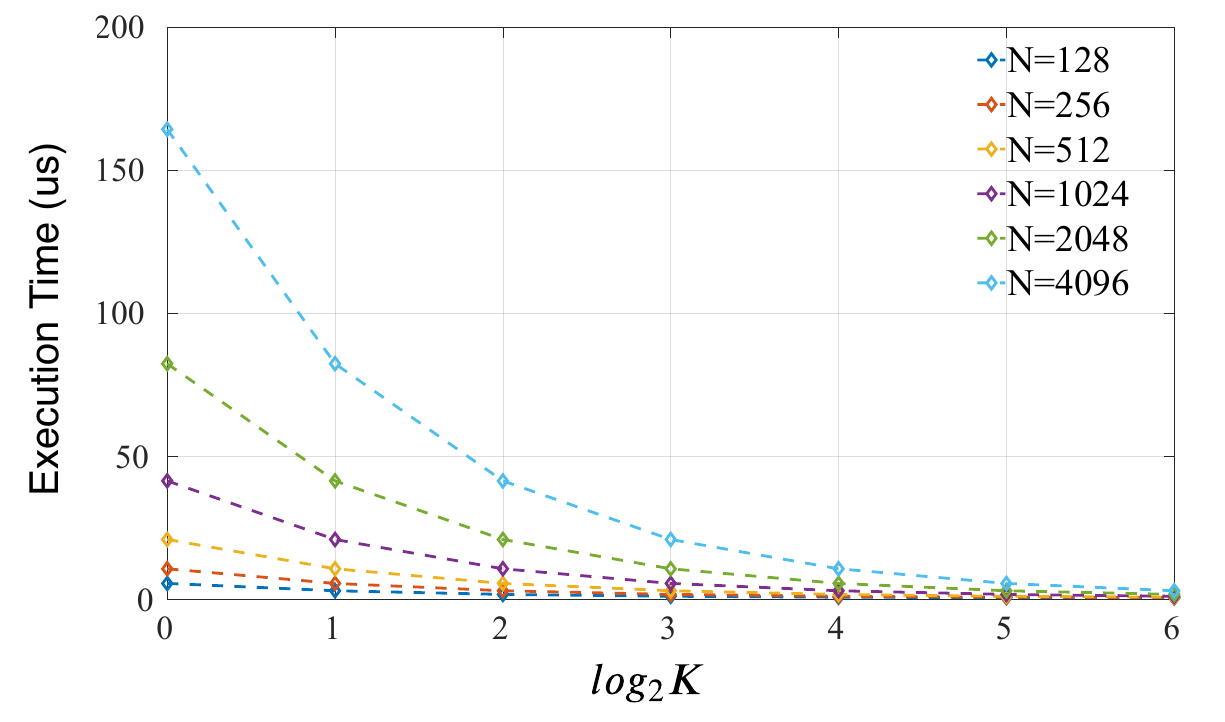}
    \caption{ }
  \end{subfigure}
    \begin{subfigure}[b]{0.47\textwidth}
    \includegraphics[width=\textwidth]{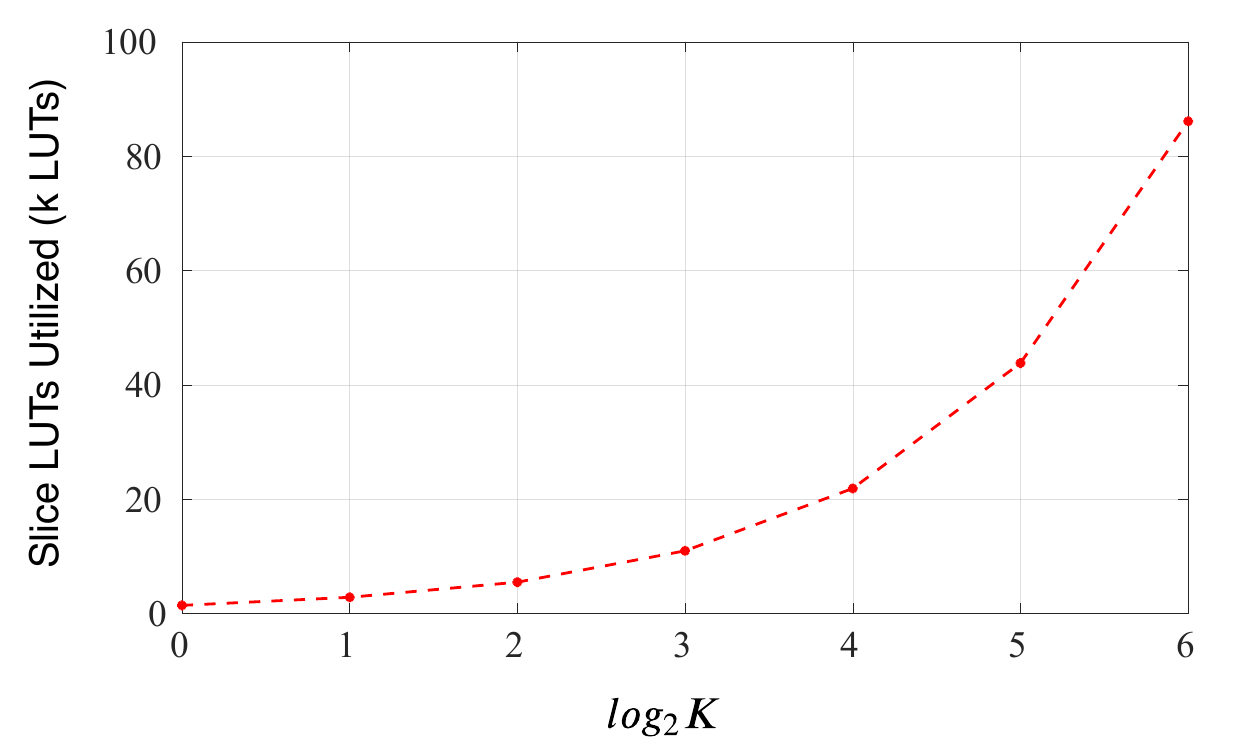}
    \caption{}
  \end{subfigure}
     \begin{subfigure}[b]{0.58\textwidth}
    \includegraphics[width=\textwidth]{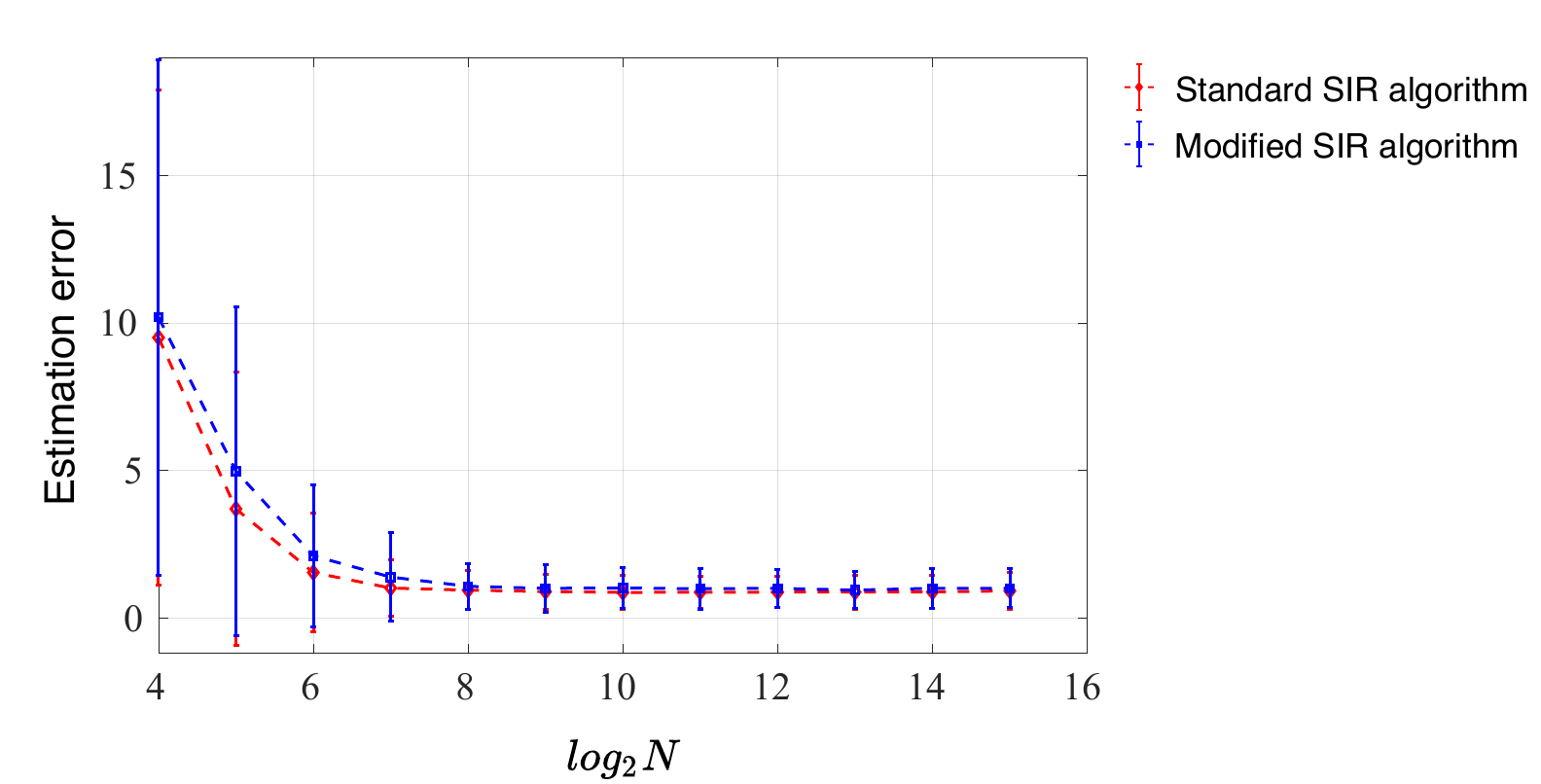}
    \caption{}
  \end{subfigure}
   
  \caption{Performance analysis of the proposed design. (a) Execution time of the proposed design as a function of the number of sub-filters $(K)$, for different number of particles $(N)$. (b) Resource utilization in terms of the number of slice LUTs used as a function of the number of sub-filters $(K)$. (c) Estimation error as a function of the number of particles $(N)$ for the standard SIR filter without any parallelization using Algorithm \ref{algo:sir_filter} and the modified SIR filter with parallelization using Algorithm \ref{algo:sir_parallel}.}
  \label{fig:ta}
\end{figure*}

 In Fig. \ref{fig:ta}(a), we show the execution time of the proposed architecture as a function of the number of sub-filters ($K$) for different $N$. As expected, the execution time increases with the number of particles $(N)$.  In many applications, for example, in biomedical signal processing, the state space dimension is very high \cite{LMiao}. Consequently, large number of particles are needed to achieve satisfactory performance. In such cases, the computation time often becomes unrealistic. Introducing parallelization in the design by using more sub-filters $(K)$ brings down the execution time significantly, as shown in Fig. \ref{fig:ta}(a). The latency used for the plot is $\tau=50$ cycles. However, the reduction in execution time by increasing $K$ comes at the cost of added hardware, which can be inferred from Fig. \ref{fig:ta}(b). Thus, there is a trade-off between the speed and the hardware utilized. For instance, using a single sub-filter, and no parallelization uses a mere $1.4$k $(1\%)$ LUTs to process 256 particles, and the time taken for SIR operations is around $1075$ clock cycles. On the other hand, with a design consisting $8$ sub-filters, it takes only $178$ clock cycles for SIR operations, but utilizes $11$k $(8\%)$ LUTs. Thus, there is a trade-off between speed and hardware used. The given FPGA resources limit the total number of sub-filters that can be accommodated on an FPGA, thus limiting the maximum achievable speed. 

\subsection{Estimation accuracy}
\label{sec:estimation_accuracy}
We analyzed the estimation accuracy for the 2D source localization problem as a function of the number of particles $(N)$ for the standard and modified SIR algorithms. The estimation error gives the error between the actual source location and the estimated source location given by:
\begin{equation}
Error = \sqrt{(pos_{x} - x)^{2}+(pos_{y} - y)^{2}}    
\end{equation}
where, $pos_{x}$ and $pos_{y}$ denote the estimated position of the source obtained from the PF algorithm, in 2D Cartesian co-ordinate system. $x$ and $y$ denote the true position of the source in the $2D$ arena.
\par The algorithm for the standard SIR filter is presented in Section. \ref{sec:PF_bg} and has no parallelization incorporated. The modified SIR algorithm implements parallelization by utilizing $K$ sub-filters working concurrently to reduce the execution time and is introduced in Section. \ref{sec:algo_mod}. The estimation errors presented in Fig. \ref{fig:ta}(c) are the average errors in $1000$ runs over $250$ time-steps.
It is inferred that there is no significant difference in the estimation error between the standard and the modified SIR algorithm. Additionally, the modified algorithm achieves lower execution time and allows the parallel computation of PFs. Further, it is noted that by scaling the number of particles, the estimation accuracy improves as the error decreases.

\subsection{Choice of number of sub-filters $K$}
Choice of the number of sub-filters $(K)$ used in the design depends on several factors, such as the number of particles $(N)$, the clock frequency of the design $(f_{clk})$, and the observation sampling rate $(f_{s})$ of the measurement samples. The sampling rate gives the rate at which new input measurements can be processed. $N$ is chosen depending on the application for which the particle filter is applied. $f_{clk}$ is selected based on the maximum frequency supported by the design. The sampling rate is related to the execution time $(T_{SIR})$ of the filter as:
\begin{equation*}
f_{s} = 1/T_{SIR} = \frac{f_{clk}}{(4N/K+\tau)}
\end{equation*}
where, $f_{clk} = 1/T_{clk}$.
Thus, for a specified measurement sampling rate $(f_{s})$, clock frequency of the design $(f_{clk})$, and the number of particles $(N)$, we can determine the number of sub-filters $(K)$ needed from the above equation. For instance, in our application, we use $256$ particles because the error curve levels off at $N=256$  (cf. Fig. \ref{fig:ta}(c)), and there is no improvement in the estimation error by further increasing $N$. Thus, to achieve a sampling rate of $f_{s}=562$ kHz, with $256$ particles and clock frequency $f_{clk}=100$ Mhz, we utilize $K=8$ sub-filters. The maximum number of sub-filters that can be used in the design depends on the resources of the given FPGA.

\begin{table*}[h]
\caption{Performance summary and comparison with state-of-the-art particle filter FPGA implementation schemes.}
\label{tab:comp}

\centering
\begin{tabular}{|M{0.16\textwidth}|M{0.13\textwidth}|M{0.13\textwidth}|M{0.13\textwidth}|M{0.13\textwidth}|M{0.13\textwidth}|}
\hline \\ [-1em]
\text{Reference} & \text{Athalye et al. \cite{Athalye}}  & \text{Ye and Zhang \cite{Ye}} & \text{Sileshi et al. \cite{Sileshi}} & \text{Velmurugan \cite{Velmurugan}} & \text{This Work} \\  \hline \\ [-1em] 
\text{ Application} & \text{Tracking} & \text{Tracking} & \text{Localization} & \text{Tracking}  & \text{Localization} \\ \hline \\ [-1em]
\text{ FPGA Device} & Xilinx Virtex II pro  & Xilinx Virtex-5 & Xilinx Kintex-7 & Xilinx Virtex II pro & Xilinx Artix-7 \\ \hline \\ [-1em]
\text{Number of particles (N)} & $2,048$  & $1,024$ & $1,024$ & $1,000$ & $1,024$ \\ \hline \\ [-1em] 
\text{Slice Registers} & $4,392$ & $13,692$ & $1,462$  & $17,418$ & $12,356$\\ \hline \\ [-1em]
\text{Slice LUTs} & $3,848$ & $7,379$ & $19,116$  & $30,900$ & $10,973$ \\ \hline \\ [-1em]
\text{Clock Frequency} & $100$ MHz  & $100$ MHz & $100$ MHz & $100$ MHz & $100$ MHz\\ \hline \\ [-1em]
\text{Execution Time ($\mu$s)} &  $60.24$ & $21.74$ & $55.37$  & $33.33$  & $\bm{5.62^{*}}$\\ \hline
\text{Sampling Rate (kHz)} &  $16$ & $46$ & $18$ & $30$ & $\bm{178^{*}}$ \\ \hline
\end{tabular}
\begin{tablenotes}
      \small
      \item $^{*}$$K=8$ sub-filters are used for the calculation.
    \end{tablenotes}
\end{table*}

\subsection{Comparison with state-of-the-art FPGA implementations} 
A comparison of our design with state-of-the-art FPGA implementation schemes is provided in Table \ref{tab:comp}. To obtain a valid assessment with other works, we have used $N=1,024$ particles (although $256$ particles are sufficient for our application as error curve levels off at $N=256$  (cf. Fig. \ref{fig:ta}(c)) and $K=8$ sub-filters for the comparison. Most of the existing implementation schemes in literature implement the standard SIR algorithm (cf. Algorithm \ref{algo:sir_filter}), which lacks parallelization. Moreover, their architectures are not scalable to process a large number of particles at a high sampling rate, as the execution time is proportional to the number of particles. Also, the re-sampling step is a major computational bottleneck, as it is inherently not parallelizable. In this work, we propose a modification to the existing algorithm that overcomes this computation bottleneck of the PF algorithm and makes the high-speed implementation possible. We introduce an additional particle routing step (cf. Algorithm \ref{algo:sir_parallel}) allowing for parallel re-sampling. We develop a PF architecture based on the modified algorithm incorporating parallelization and pipelining design strategies to reduce the execution time. Our design achieves very high input sampling rates, even for a large number of particles, by scaling the number of sub-filters $K$. 
\par The first hardware architecture for implementing PFs on an FPGA was provided by  Athalye et al. \cite{Athalye}, applied to a tracking problem. Their architecture is generic and does not incorporate any parallelization in the design. Thus, their architecture suffers from a low sampling rate of about $16$ kHz for $2048$ particles, which is approximately $11$ times lower than the sampling rate of our design. However, owing to the non-parallel architecture, the resource consumption of their design ($4.3$k registers and $3.8$k LUTs) is relatively low.  Another state-of-the-art system was presented in \cite{Ye}. The authors implemented an SIR filter on the Xilinx Virtex-5 FPGA platform for bearings-only tracking application and achieved a sampling rate of $46$ kHz for $1024$ particles. Regarding its hardware utilization, it uses $13.6$k registers and $7.3$k LUTs, which are comparable to those of our design; however, their sampling rate is four times lower than that of our system. Sileshi et al. \cite{Sileshi} proposed two methods for implementing PFs on hardware. The first method was a hardware/software (HW/SW) co-design approach, where the software components were implemented using an embedded processor (MicroBlaze). The hardware part was based on a PF hardware acceleration module on an FPGA. 
This HW/SW co-design approach has a low sampling rate of about $1$ kHz due to communication overhead between the MicroBlaze soft processor and the hardware acceleration module. Further, speedup of the design by utilizing a large number of parallel particle processors is limited by the number of bus interfaces available in the soft-core processor (MicroBlaze). Thus, to improve the sampling rate, they proposed a second approach which is entirely a hardware design. However, their architecture does not support parallel processing and achieves a low sampling rate of about $18$ kHz, whereas our system can sample at $178$ kHz for processing the same $1024$ particles. Their full hardware system utilizes $1.4$k registers and $19$k LUTs. Velmurugan \cite{Velmurugan} proposed a fully digital PF FPGA implementation for tracking application, without any parallelization in the design. They used a high-level Xilinx system generator tool to generate the VHDL code for deployment on a Xilinx FPGA from Simulink models or MATLAB code. Their design is not optimized in terms of hardware utilization as they use a high-level abstraction tool and lack flexibility to fine-tune the design. On the other hand, our design is completely hand-coded in Verilog and provides granular control to tweak the design parameters and provides flexibility for easy integration of the design in numerous PF applications. They achieve a sampling rate of about $30$ kHz for $1000$ particles, which is six times lower than that of our design. Further, their resource consumption is relatively high ($17.4$k registers and $30.9$k LUTs) as they use high-level abstraction tools for implementation. 
\par Our system has a comparable resource utilization ($12.3$k registers and $10.9$k LUTs for $8$ sub-filters) with a low execution time of about $5.62$ $\mu$s and can achieve a very high input sampling rate of about $178$ kHz compared to other designs. Our design can be used in real-time applications due to the low execution time. Further, to achieve a high sampling rate even with a large number of particles, more sub-filters can be used, as shown in Fig. \ref{fig:ta}(a). However, this comes at the cost of added hardware. On the other hand, the resource utilization of our system can go as low as $1.7$k registers and $1.4$k LUTs using a single sub-filter (cf. Table \ref{tab:ru}) for applications that have stringent resource constraints, but at the cost of increased execution time (cf. Fig. \ref{fig:ta}(a)). 

\subsection{Experimental results} 
\begin{figure*}[h]
\centering
\includegraphics[width=12cm]{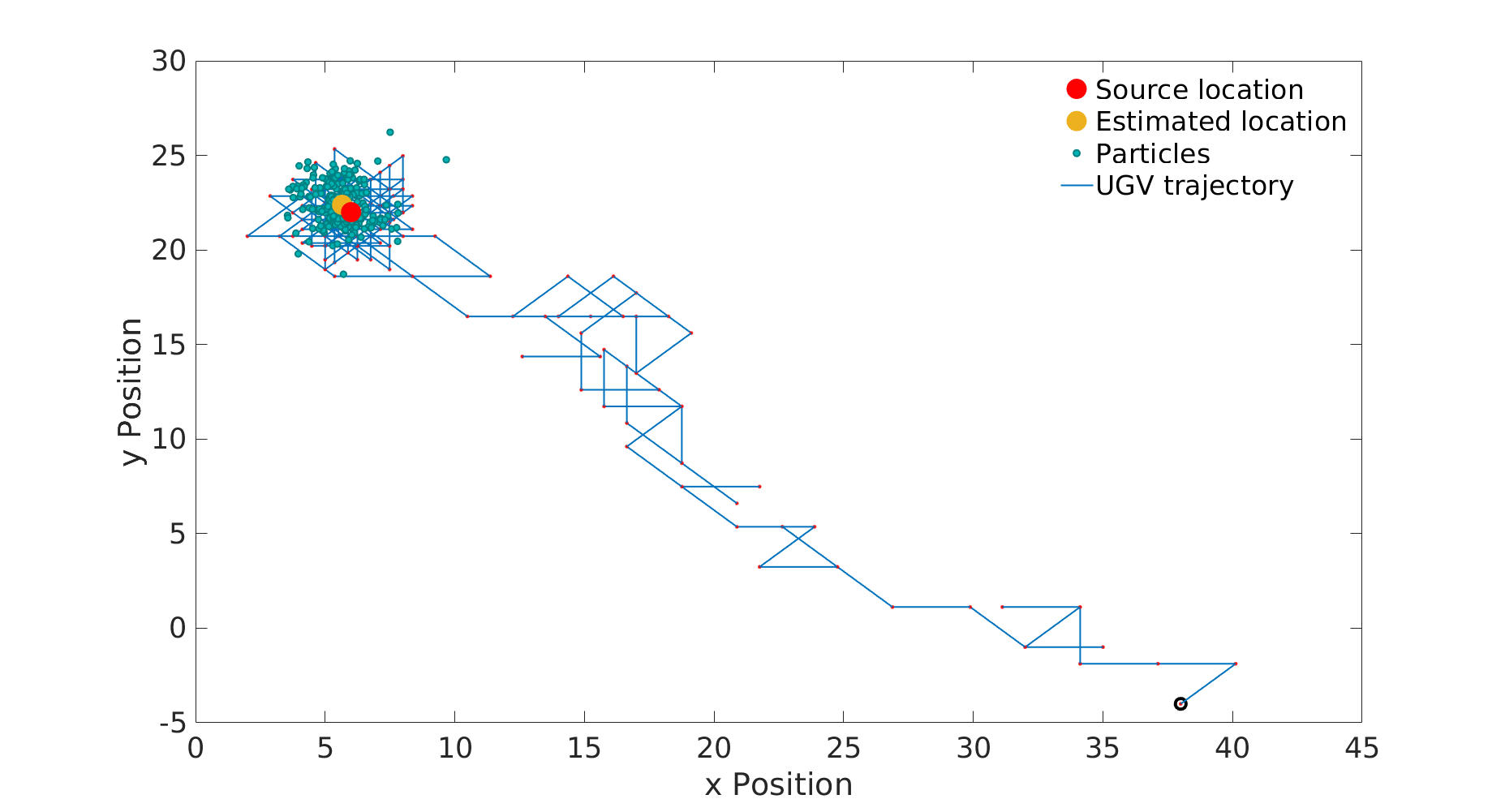}
\centering\caption{2D source localization experimental result. The source is positioned at $[6,22]$ marked by a 'red' circular dot. At the start, the UGV is positioned at $[38,-4]$. The model is run over 250 time-steps for 256 particles, and the UGV traverses towards the source based on sensor measurements. The final source estimate $(pos_{t})$ obtained by the PF algorithm is marked by a 'yellow' circular dot and has an estimation error of $0.5$.}
\label{fig:sl_2d}
\end{figure*}

\begin{figure*}[h]
\centering
\includegraphics[width=8cm]{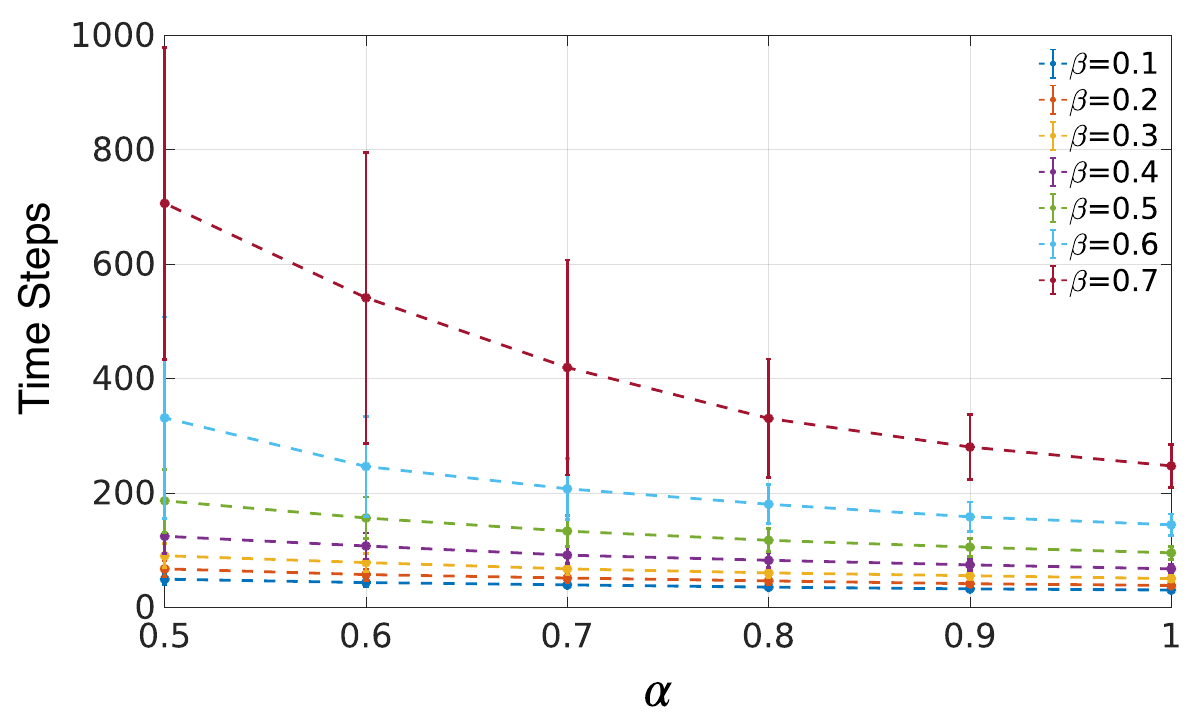}
\centering\caption{Variation in the number of time-steps required to localize the source as a function of $\alpha$ and $\beta$.}
\label{fig:ts}
\end{figure*}

The experimental result for the 2-dimensional source localization is shown in Fig. \ref{fig:sl_2d}. State in this 2D model is 2-dimensional and incorporates position in x and y directions. The input to the design is binary measurements from a set of $8$ photodiodes and the instantaneous position of the UGV. The inputs are sampled and processed by the PF system over $250$ time-steps on an FPGA to estimate the source location. We consider the probabilities $\alpha$ and $\beta$ to be $0.8$ and $0.6$, respectively. It can be seen that the algorithm is robust enough to localize the source even with a noise probability of $0.6$. However, with an increase in noise probability $(\beta)$, the number of time-steps or iterations required to localize the source also increases, as shown in Fig. \ref{fig:ts}. The source is considered to be localized if the estimation error is less than the predetermined threshold, which is $2.5$ in our case. The time-steps shown in Fig. \ref{fig:ts} are the average time required to localize the source over 500 runs. The entire design was coded in Verilog HDL, and the design was implemented on an FPGA. 

\begin{figure*}[h!]
\centering
\includegraphics[width=13.5cm]{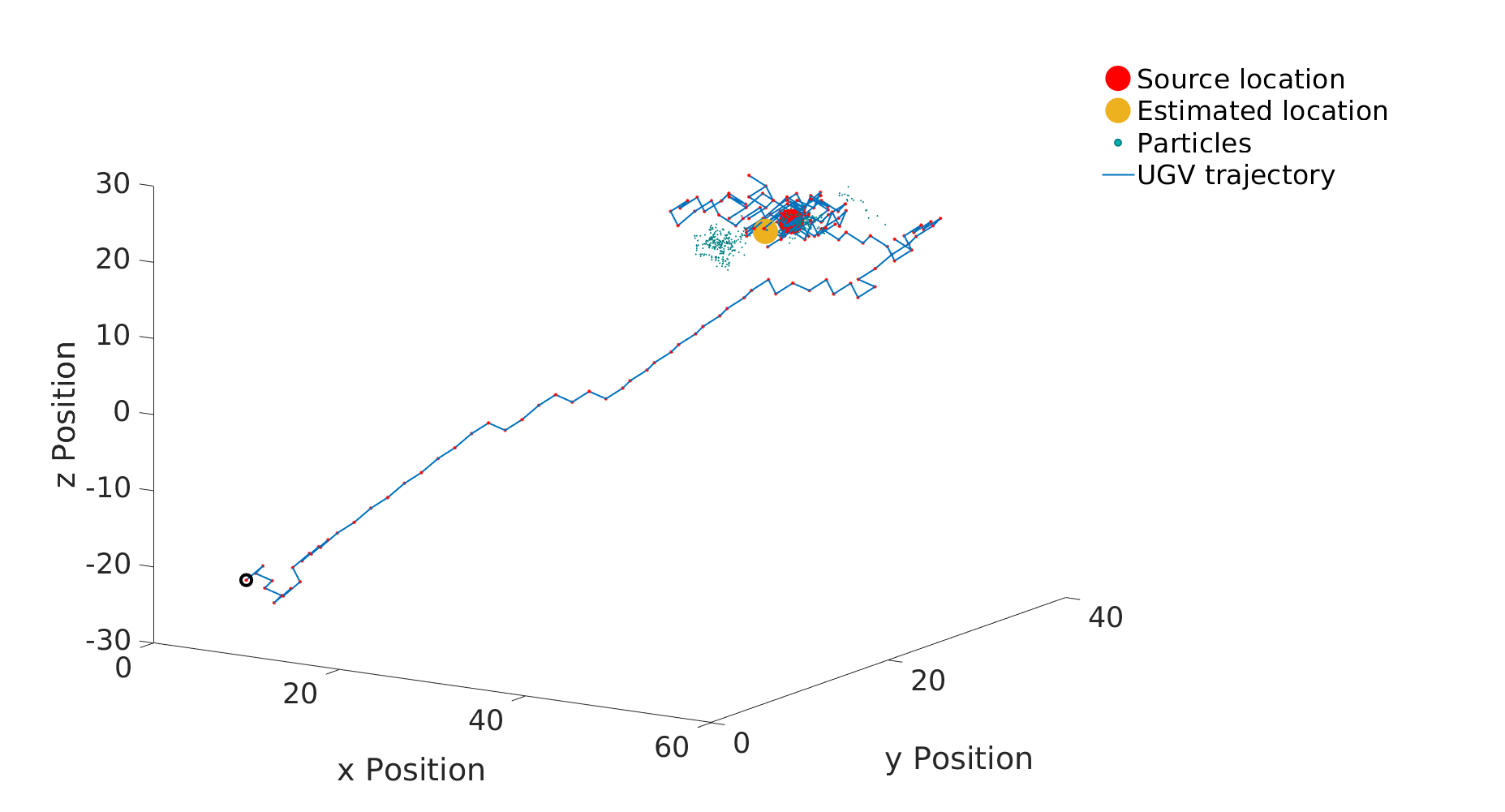}
\centering\caption{3D source localization experimental result. The source is positioned at $ [40,30 ,20]$, and the initial position of the UGV is  $[10,0,-20]$. The model runs over 350 time-steps for 512 particles. Here, the UAV traverses in three dimensions to move towards the source. The error between the source and the estimated location is $2.2$.}
\label{fig:sl_3d}
\end{figure*} 

\par 
We converted all variables from a floating-point to a fixed point representation for the implementation on an FPGA. We have used a $16$-bit fixed-point representation for particles and their associated weights. All bearing-related information, such as the angle of the UGV and the angle of particles used in the importance block, is represented by a $12$-bit fixed-point representation. Further, the indices of the replicated and the discarded particles are integers and are represented using $log_{2}(M) = 5$ bits. The output estimate of the source location $(pos_{t})$ is represented using a $16$-bit representation.
  $N=256$ particles were used for processing. $K=8$ sub-filters were used in the design with $M=32$ particles processed within each sub-filter. $M/2=16$ particles were exchanged between the sub-filters after the completion of every iteration as a part of particle routing operation. The time taken to complete SIR operations for $N=256$ and $K=8$ is $178$ clock cycles. Using a clock frequency of $100$ MHz, the speed at which we can process new samples is around $562$ kHz, and the execution time for SIR operation is $1.78$ $\mu$s. This high sampling rate enables us to use the proposed hardware architecture in various real-time applications.

\par Further, we show that the 2D source localization problem can be extended to 3D, and we have modelled it in software using MATLAB. This 3D model incorporates position along the x, y, and z directions. Here, an Unmanned Aerial Vehicle (UAV) can be utilized to localize the source. As compared to $8$ sensors used in 2D localization, here we utilize $16$ sensors for scanning the entire 3D space. We consider $\alpha=0.8$ and $\beta=0.4$, and the model was run over $350$ time-steps for $512$ particles to localize the source. The result is presented in Fig. \ref{fig:sl_3d}. The estimation error between the actual source location and the estimated source location in the 3D arena is given by:

\begin{equation}
Error = \sqrt{(pos_{x} - x)^{2}+(pos_{y} - y)^{2}+(pos_{z} - z)^{2}}    
\end{equation}
where, $pos_{x}$, $pos_{y}$ and $pos_{z}$ denote the estimated position of the source obtained from PF algorithm, in 3D Cartesian co-ordinate system. $x$, $y$ and $z$ represent the true position of the source in $3D$ arena.

\section{Conclusions and Outlook}
\label{sec:conc}
In this paper, we presented an architecture for hardware realization of PFs, particularly sampling, importance, and re-sampling filters, on an FPGA. PFs perform better than traditional Kalman filters in non-linear and non-Gaussian settings. Interesting insights into the advantages of PFs, performance comparison, and trade-offs of PFs over other non-PF solutions are provided by \cite{Ababsa,Ko}.
However, PFs are computationally very demanding and take a significant amount of time to process a large number of particles; hence, PFs are seldom used for real-time applications. In our architecture, we try to address this issue by exploiting parallelization and pipelining design techniques to reduce the overall execution time, thus making the real-time implementation of PFs feasible. However, a major bottleneck in high-speed parallel implementation of the SIR filter is the re-sampling step, as it is inherently not parallelizable and cannot be pipelined with other operations. In this regard, we modified the standard SIR filter to make it parallelizable. The modified algorithm has an additional particle routing step and utilizes several sub-filters working concurrently and performing SIR operations independently on particles to reduce the overall execution time. The architecture presented is massively parallel and scalable to process a large number of particles and has low design complexity. 
\par A performance assessment in terms of the resources utilized on an FPGA, execution time, and estimation accuracy is presented. We also compared the estimation error of the modified SIR algorithm with that of the standard SIR algorithm and noted that there is no significant difference in the estimation error. The proposed architecture has a total execution time of about $5.62$ $\mu$s (i.e., a sampling rate of $178$ kHz) by utilizing $8$ sub-filters for processing $N=1024$ particles. We compared our design with state-of-the-art FPGA implementation schemes and found that our design outperforms other implementation schemes in terms of execution time. The low execution time (i.e., high input sampling rate) makes our architecture ideal for real-time applications.
\par The proposed PF architecture is not limited to a particular application and can be used for other applications by modifying the importance block of the sub-filter. The sampling and re-sampling block designs are generic and can be used for any application. In this work, we validated our PF architecture for the source localization problem to estimate the position of a source based on received sensor measurements. Our PF implementation is robust to noise and can predict the source position even with a high noise probability. Experimental results show the estimated source location with respect to the actual location for 2D and 3D settings and demonstrate the effectiveness of the proposed algorithm. 
\par In recent times, there has been an increase in the utilization of UGVs in several instances, such as disaster relief and military applications, due to reduced human involvement and the ability to carry out the task remotely. The proposed source localization framework using PFs can autonomously navigate and localize the source of interest without any human intervention, which would be very helpful in missions wherein there is an imminent threat involved, such as locating chemical, biological or radiative sources in an unknown environment. Further, the proposed PF framework and its hardware realization would be useful for the signal processing community for solving various state estimation problems such as tracking, navigation, and positioning in real-time.

\bibliographystyle{IEEEtran}
\bibliography{particle_filters.bib}

\end{document}